\documentclass[letterpaper,11pt]{article}

\usepackage[dvipdfmx]{graphicx}
\usepackage{amssymb} 
\usepackage{amsmath}

\usepackage{comment}
\usepackage{color} 

\usepackage{algorithmic,algorithm} 

\newtheorem{theorem}{Theorem}

\title{Rendezvous and Merging for Two Metamorphic Robotic Systems without Global Compass} 
\author{Ryonosuke Yamada\footnote{Graduate School of ISEE, Kyushu University, Japan.}, 
Tomoyuki Usami\footnote{Graduate School of ISEE, Kyushu University, Japan.}, 
and Yukiko Yamauchi\footnote{Faculty of ISEE, Kyushu University, Japan}} 

\date{}

\begin{document}
\maketitle

{\noindent{\bf Abstract. }} 
A \emph{metamorphic robotic system (MRS)} consists of 
anonymous \emph{modules}, each of which autonomously moves in the 2D square grid 
by sliding and rotation with keeping connectivity among the modules. 
Existing literature considers distributed coordination among modules 
so that they collectively form a single MRS. 
In this paper, we consider distributed coordination for two MRSs. 
We first present a rendezvous algorithm that makes the two MRSs gather 
so that each module can observe all the other modules. 
Then, we present a merge algorithm that makes the two MRSs assemble and establish 
connectivity after rendezvous is finished. 
These two algorithms assume that each MRS consists of five modules, 
that do not have a common coordinate system. 
Finally, we show that five modules for each MRS is necessary 
to solve the rendezvous problem. 
To the best of our knowledge, our result is the first result on 
distributed coordination of multiple MRSs. 


\section{Introduction}
\label{sec:intro}

\emph{The rendezvous problem} is the simplest agreement problem in 
distributed coordination of autonomous mobile computing entities. 
The problem requires two mobile computing entities to gather at some position, 
which is not given in advance. 
The problem has been considered for mobile agents~\cite{KSSK03,DPP23} 
and mobile robots~\cite{SY99} in a graph and the 2D Euclidean space. 
Most existing literature presents impossibility results, which conversely 
show necessary assumptions and equipment of the mobile computing entities. 
For example, two mobile robots can solve the rendezvous problem in the fully-synchronous model, 
while they cannot in the semi-synchronous and asynchronous models~\cite{SY99}. 
However, when each robot is equipped with a light, 
that can take a constant number of colors, 
the rendezvous problem is solvable in the semi-synchronous and asynchronous models~\cite{DFPSY16}. 

\begin{figure}[t]
 \begin{tabular}{cc}
  \begin{minipage}[t]{0.45\hsize}
   \centering
   \includegraphics[width=3.5cm]{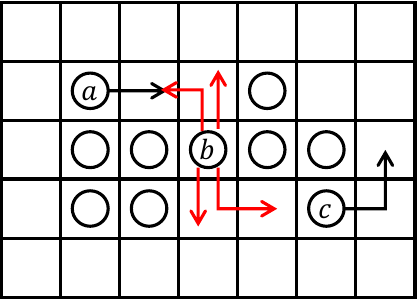} 
   \caption{Movements and connectivity} 
   \label{fig:intro-1} 
  \end{minipage} &
  \begin{minipage}[t]{0.45\hsize}
   \centering
   \includegraphics[width=3.5cm]{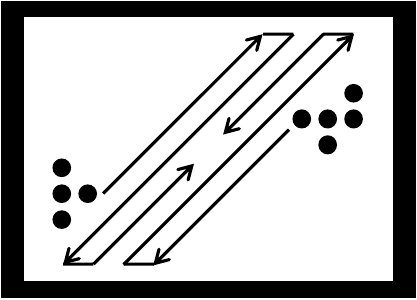}
   \caption{Rendezvous of to MRSs} 
   \label{fig:intro-2} 
      \end{minipage} 
    \end{tabular}
  \end{figure}

In this paper, we investigate the rendezvous problem for 
the \emph{metamorphic robotic systems (MRSs)}. 
An MRS consists of a collection of autonomous mobile computing entities called 
\emph{modules}, each of which autonomously moves in the 2D square gird. 
Each cell of the square grid can accept at most one module at each time step. 
Modules are \emph{anonymous} (indistinguishable) 
and each module can perform a \emph{rotation} or a \emph{sliding} at one time. 
At each time step, more than one modules can perform rotations and slidings 
as long as there is no collision and the modules keep their connectivity. 
We consider a graph over the models, where the vertex set is the the set of modules and 
two modules are connected by an undirected edge when they are placed at side-adjacent cells. 
Then, an MRS is \emph{connected} if the graph over the modules is connected. 
Figure~\ref{fig:intro-1} shows an example, 
where module $a$ can perform a sliding and module $c$ can perform a rotation.  
However, module $b$ cannot perform the rotations or slidings shown by the red arrows 
because these movements break the connectivity of the MRS. 

A variety of problems have been considered for a single MRS. 
The \emph{reconfiguration problem} requires an MRS to change its initial shape 
to a given target shape~\cite{DSY04a}. 
Dumitrescu et al. showed that any horizontally convex shape can be transformed to 
another horizontally convex shape via a canonical chain shape. 
The \emph{locomotion problem} requires an MRS to move toward a given direction. 
Dumitrescu et al. presented a shape that achieves fastest locomotion 
to the horizontal direction and another shape for the orthogonal direction~\cite{DSY04b}.
The \emph{search problem} requires an MRS to find a target in a field, 
which is a rectangular field surrounded by walls.  
Doi et al. pointed out that the necessary and sufficient number of modules 
to solve a problem represents the complexity of the problem~\cite{DYKY22}. 
That is, as the number of modules increases, the number of shapes (i.e., states) 
of an MRS increases and 
the MRS is expected to be able to solve more complicated problems. 
The authors also investigated the effect of common knowledge of the coordinate systems. 
The modules are equipped with the \emph{global compass} when they agree on 
the north, south, east, and west in the 2D square grid. 
They showed that three modules are necessary and sufficient for the search problem 
if the modules are equipped with global compass,  
otherwise five modules are necessary and sufficient. 
The \emph{evacuation problem} requires an MRS to get out of a field through 
an exit placed on the boundary of the field. 
Nakamura et al. showed that two modules are necessary and sufficient 
if the modules are equipped with global compass,  
otherwise four modules are necessary and sufficient for a rectangular field~\cite{NKY22}.  
Existing results consider distributed coordination among the modules 
so that the modules collectively form a single MRS. 

In this paper, we consider distributed coordination among MRSs. 
We first consider the rendezvous problem for two MRSs placed in a square field. 
The goal of the rendezvous problem is to gather the two MRSs 
so that each module can observe all modules of the two MRSs in its 
constant visibility. 
When the modules are equipped with the global compass, 
the two MRSs can solve the rendezvous problem in the same approach as 
the evacuation problem~\cite{NKY22}.
That is, the two MRSs first move to some direction until it reaches a wall, 
and then move toward the south west corner. 
However, this algorithm cannot solve the rendezvous problem when the modules 
are not equipped with the global compass. 
In the proposed algorithm, we adopt an orthogonal ``path'' in the field 
and make the two MRSs move back and forth on the path. 
Thus, they can finish rendezvous in finite time. 
See Figure~\ref{fig:intro-2} as an example. 
However, an MRS may follow a mirror image of the path due to the lack of 
the global compass. 
We will show that an MRS can return to the correct path if they are not moving on it. 
The proposed algorithm is designed for two MRSs each of which consists of five modules 
not equipped with the global compass.  
We also show that five modules are necessary to solve the rendezvous problem 
based on the results in \cite{DYKY22}. 

We then introduce the \emph{merge problem} that requires the modules of the 
two MRSs to assemble and establish entire connectivity 
from a configuration where each module of the two MRSs can observe all the other modules. 
A simple solution for the merge problem is to make one MRS 
wait the other MRS to come closer and merge. 
The proposed merge algorithm is based on a labeling of all $18$ shapes of five modules 
and it makes the MRS $R$ with the smaller label move toward the other MRS $R'$ with the larger label. 
When the two MRS are initially in the same shape, 
each module first computes the \emph{view} of each MRS, 
which is a configuration of the modules 
observed in the local coordinate system defined for each shape of the MRS. 
Then the MRS with a smaller view moves toward the MRS with a larger view. 
Finally, we present exceptional movements for initial configurations,  
where the two MRS cannot break their tie by their shapes or views.  
To the beset of our knowledge, this is the first result on 
distributed coordination for two MRSs. 
We believe our results open up new vistas for the distributed coordination theory 
of the MRS model. 

\noindent{\bf Related work.} 
The conventional evaluation criteria for an MRS is 
the total number of movements by modules and execution time. 
Michail et al. considered reconfiguration under different connectivity requirements and 
different set of movements~\cite{MSS19}. 
They showed that there exists two shapes that cannot be reconfigured to each other 
by rotation only. 
Then they showed if connectivity is not required, 
reconfiguration by rotation is in P when a pair of modules are given from the outside, 
otherwise it is in PSPACE. 
Then, they showed that reconfiguration with connectivity is solvable for any two shapes 
via a canonical chain shape by rotation and sliding. 
They further showed that this simple reconfiguration requires $\Theta(n^2)$ moves, 
which can be reduced to $O(n)$ parallel time by a pipelining strategy. 
Almethen et al. introduced a new movement called \emph{line move} 
that shifts all modules forming a vertical or horizontal line by one cell~\cite{AMP20}. 
They showed that any connected shape can be reconfigured to another 
connected shape in $O(n \log n)$ time by rotations, slidings, and line moves 
when the connectivity is not required and in $O(n \sqrt{n})$ time when 
the connectivity is required. 

After Doi et al. introduced the notion of necessary and sufficient number of 
modules to solve a problem~\cite{DYKY22}, 
several papers considered extensions of these results to more general field. 
Nakamura et la. showed that their evacuation algorithms for 
a rectangular field in the 2D square grid can be extended to 
a maze and a convex field~\cite{NKY22}. 
That is, two modules are necessary and sufficient if the modules are equipped 
with the global compass, 
otherwise four modules are necessary and sufficient. 
Yamada and Yamauchi considered the search problem in a 3D cubic grid~\cite{YY22}. 
In the 3D space, the global compass consists of agreement of 
the directions and orientations of the $x$-axis, $y$-axis, and $z$-axis. 
The authors considered an intermediate model, where the modules agree on 
the direction and orientation of the vertical axis. 
The authors showed if modules are equipped with the global compass, 
three modules are necessary and sufficient, otherwise  
five modules are necessary and sufficient. 
Then, they showed that in the intermediate model, 
four modules are necessary and sufficient. 

{\bf Organization.} 
Section~\ref{sec:prel} provides the MRS model, 
and definitions of the rendezvous problem and the merge problem. 
We present a rendezvous algorithm in Section~\ref{sec:rendezvous} and 
a merge algorithm in Section~\ref{sec:merge} for two MRSs 
each of which consists of five modules not equipped with the global compass.  
Section~\ref{sec:impossibility} shows that 
five modules are necessary and sufficient in our settings. 
We conclude the paper and present future directions in Section~\ref{sec:concl}.  

\section{Preliminary}
\label{sec:prel}

\subsection{Metamorphic robotic system}

A \emph{metamorphic robotic system} consists of a set of 
anonymous \emph{modules} that autonomously moves in a 2D square grid. 
Let $R = \{m_1, m_2, \ldots, m_n\}$ be an MRS that consists of $n$ modules. 
We call $n$ the \emph{size} of $R$. 
We use $m_i$ just for notation. 
When we consider more than one MRSs $R_1, R_2, \ldots$, 
each MRS is represented by $R_i = \{ m_{i,1}, m_{i,2}, \ldots, m_{i, n_i} \}$, 
where $n_i$ is the seize of $R_i$. 
In this case, we also use $R_i$ and $m_{i,j}$ just for notation. 

\begin{figure}[t]
 \begin{tabular}{cc}
  \begin{minipage}[t]{0.55\hsize}
   \centering
   \includegraphics[width=6cm]{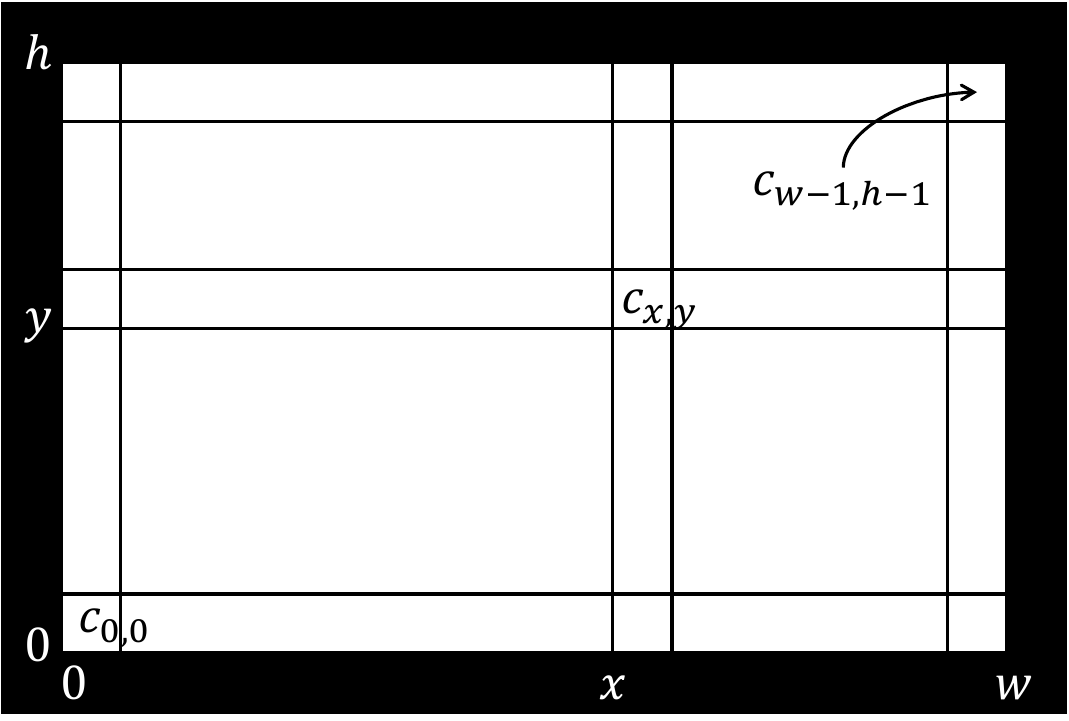} 
   \caption{Field and walls}
   \label{fig:field}
  \end{minipage} &
  \begin{minipage}[t]{0.4\hsize}
   \centering
   \includegraphics[width=3cm]{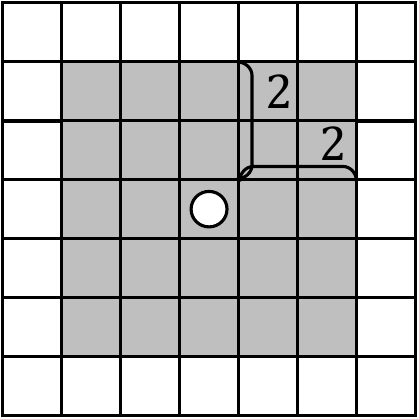} 
   \caption{Visibility when $k=2$} 
   \label{fig:visibility}
      \end{minipage} 
    \end{tabular}
  \end{figure}

We consider MRSs moving in a finite \emph{field} in the 2D square grid. 
The field is a rectangle of width $w$ and height $h$ with $w \neq h$. 
Without loss of generality, we assume $w > h$. 
We consider the global coordinate system, whose origin is a corner of the field. 
Then, we call the positive $y$ direction \emph{north} 
and the negative $y$ direction \emph{south}. 
Thus, the positive $x$ direction is \emph{east} 
and the negative $x$ direction is \emph{west}. 
A cell whose bottom left coordinate is $(x, y)$ is denoted by $c_{x,y}$. 
Then, the field consists of cells $c_{x,y}$ for $ x \in \{0, 1, \ldots, w-1\}$ 
and $y \in \{0, 1, \ldots, h-1\}$. 
The field is surrounded by four walls, i.e., 
the north wall $W_N = \{c_{x,y} \mid x \in \{-1, 0, 1, \ldots, w\}, y = h\}$, 
the south wall $W_S = \{c_{x,y} \mid x \in \{-1, 0, 1, \ldots, w\}, y = -1\}$, 
the east wall $W_E = \{c_{x,y} \mid x = w, y \in \{-1, 0, 1, \ldots, h\}\}$, 
and the west wall $W_W = \{c_{x,y} \mid x = -1, y \in \{-1, 0, 1, \ldots, h\}\}$. 
See Figure~\ref{fig:field}, where walls are represented by black cells.  

Cell $c_{x,y}$ is \emph{side-adjacent} to four cells, 
$c_{x-1, y}$, $c_{x, y-1}$, $c_{x+1, y}$, and $c_{x, y+1}$. 
We also say that module $m_j$ is side-adjacent to module $m_i$ 
if the two cells occupied by $m_i$ and $m_j$ are side-adjacent. 
Each cell can accept at most one module and we say a cell is \emph{empty} 
when there is no module in the cell. 
A \emph{configuration} $C_t$ at time $t$ is the set of cells 
occupied by the modules at time $t$. 
We now define the connectivity requirement for a single MRS $R = \{m_1, m_2, \ldots, m_n\}$. 
We consider the \emph{side-adjacency} graph $G_t = (R, E_t)$, 
where $\{m_i, m_j\} \in E_t$ if $m_i$ is side-adjacent to $m_j$. 
We say $R$ is \emph{connected} at time $t$ if $G_t$ is connected. 

The modules have limited visibility. 
When a module occupies a cell $c_{x,y}$, 
it can observe whether each cell in its \emph{$k$-neighbor}  
$\{c_{x+i, y+j} \mid i,j \in [-k, k] \} $
is occupied or not. 
We call the value of $k$ the \emph{visibility range} of the module. 
See Figure~\ref{fig:visibility} as an example. 
Each module uses its own \emph{local coordinate system} when they observe 
the cells in its visibility. 
We assume that each local coordinate system is right-handed, 
its origin is the center of the cell that the module occupies, 
and its axes are parallel to the rows and columns of the 2D grid. 
When all local coordinate systems have the same directions and orientation, 
we say the modules are equipped with the \emph{global compass}. 
When the modules are not equipped with the global compass, 
they can still agree on the clockwise direction. 
A \emph{state} of an MRS is its local shape. 
When the modules are not equipped with the global compass, 
a state of an MRS does not contain directions. 
For example, consider the following two configurations of an MRS of two modules: 
$C = \{c_{x,y}, c_{x+1, y}\}$ and $C' = \{c_{x,y}, c_{x, y+1}\}$. 
If the modules are equipped with the global compass, 
the state of the MRS in $C$ is different from that in $C'$. 
Otherwise, the state of the MRS in $C$ is the same as that in $C'$. 

\begin{figure}[t]
    \begin{tabular}{cc}
      \begin{minipage}[t]{0.45\hsize}
        \centering
       \includegraphics[height=2.5cm]{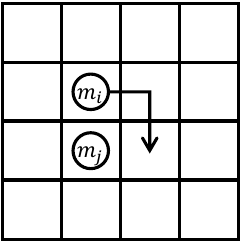}
       \caption{Rotation of $m_i$ around $m_j$.} 
       \label{fig:rotation}
      \end{minipage} &
      \begin{minipage}[t]{0.5\hsize}
        \centering
       \includegraphics[height=2.5cm]{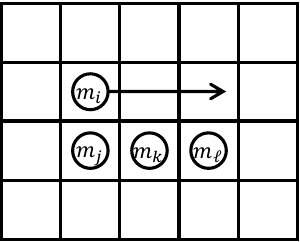}
       \caption{$2$-sliding of $m_i$ along $m_j, m_k, m_{\ell}$.} 
       \label{fig:2sliding}
      \end{minipage} 
    \end{tabular}
  \end{figure}

Each module can perform a \emph{rotation} or a \emph{sliding} at each time step. 
Each movement requires other modules that guide the movement and 
does not move during the movement, 
and it must go through empty cells. The destination cell also must be empty. 
In addition, the modules cannot go through nor enter the cells of the 
four walls surrounding the field.
\begin{itemize}
\item A rotation is a rotation by $2/\pi$ around a side-adjacent module 
that does not move simultaneously. 
See Figure~\ref{fig:rotation} as an example, where 
module $m_i$ performs a rotation around module $m_j$, 
which is side-adjacent to $m_i$ and does not move simultaneously. 
\item An \emph{$\ell$-sliding} is a vertical or horizontal movement by $\ell$-cells 
along a line of modules that do not move simultaneously. 
See Figure~\ref{fig:2sliding} as an example, where 
module $m_i$ performs $2$-sliding while 
the three modules $m_j, m_k, n_{\ell}$ along the moving track 
does not perform any movement simultaneously. 
\end{itemize}

We consider discrete time $t = 0, 1, 2, \ldots$. 
The modules are \emph{synchronous}, i.e., 
at each time instance, all modules observe its visibility, 
computes its movement, and performs the computed movement. 
The modules are \emph{uniform}, i.e., 
a module computes it movement by a common deterministic algorithm. 
The modules are \emph{oblivious}, i.e., 
when a module computes it next movement, 
the input to the common algorithm is its preceding observation. 
That is, the modules cannot use past observations nor computation. 
Let $B_t$ be the set of \emph{backbone modules} that do not move at time $t$.  
At each time instance $t$, the modules of a single MRS $R$ must keep the following conditions: 
\begin{enumerate}
\item The resulting state of $R$ is connected. 
\item $B_t$ is connected. 
\item Moving trajectories of the modules do not overlap.  
\end{enumerate}
Hence, more than one modules can move at each time instance as long as the above condition is satisfied. 

When we consider multiple MRSs $R_1, R_2, \ldots$, 
each MRS must satisfy the first two conditions on connectivity, 
and moving trajectories of all modules must not overlap. 

The movements of modules generate the evolution of the MRS(s). 
An \emph{execution} from an initial configuration $C_0$ is a sequence of configurations 
$C_0, C_1, C_2, \ldots$, 
where $C_{t+1}$ is obtained from $C_t$ by the movements of modules at time $t$.
We say an algorithm solves a problem from an initial configuration $C_0$ 
if the execution starting from $C_0$ reaches configuration $C_t$ in a finite time 
and $C_t$ satisfies the problem requirement and all modules do not move after $t$.

\subsection{Problem definitions}

The \emph{rendezvous problem} requires two MRSs initially placed in the field  
to gather so that 
each module of the two MRSs observe all the modules. 

We say two MRSs $R_1$ and $R_2$ \emph{merge} into a single MRS 
when the side adjacency graph with the vertex set $R_1 \cup R_2$ is connected. 
The \emph{merge problem} requires the two MRSs to merge into a single MRS. 
In this paper, for the merge problem, we consider initial configurations 
where each module can observe all the modules of the two MRSs.

\section{Rendezvous algorithm}
\label{sec:rendezvous}

\begin{figure}[t]
    \begin{tabular}{cc}
      \begin{minipage}[t]{0.45\hsize}
        \centering
       \includegraphics[width=4cm]{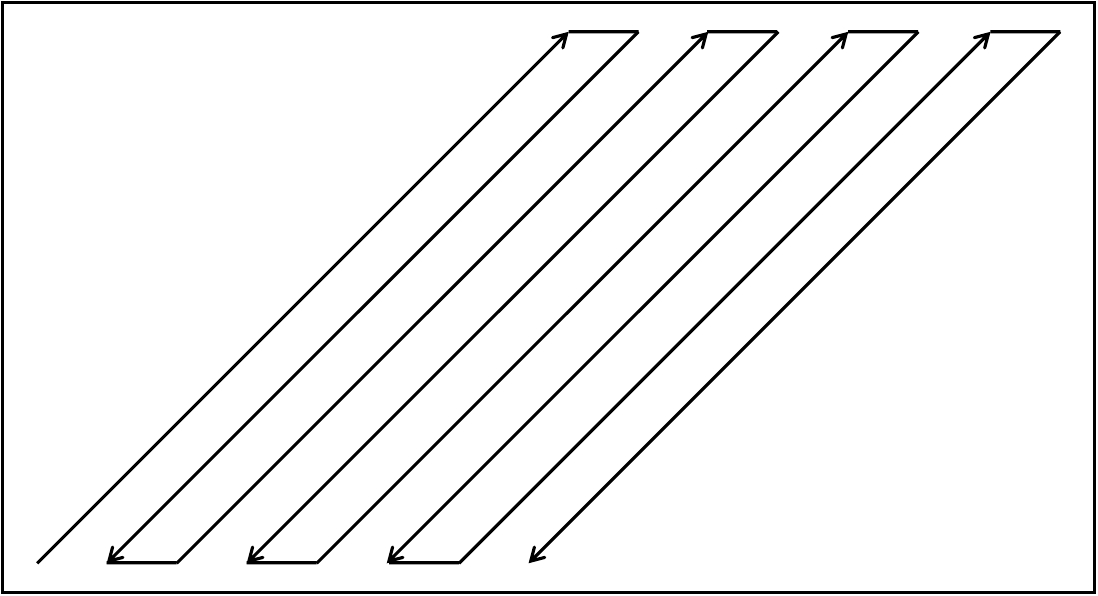} \\ 
       $(a)$
      \end{minipage} &
      \begin{minipage}[t]{0.45\hsize}
        \centering
       \includegraphics[width=4cm]{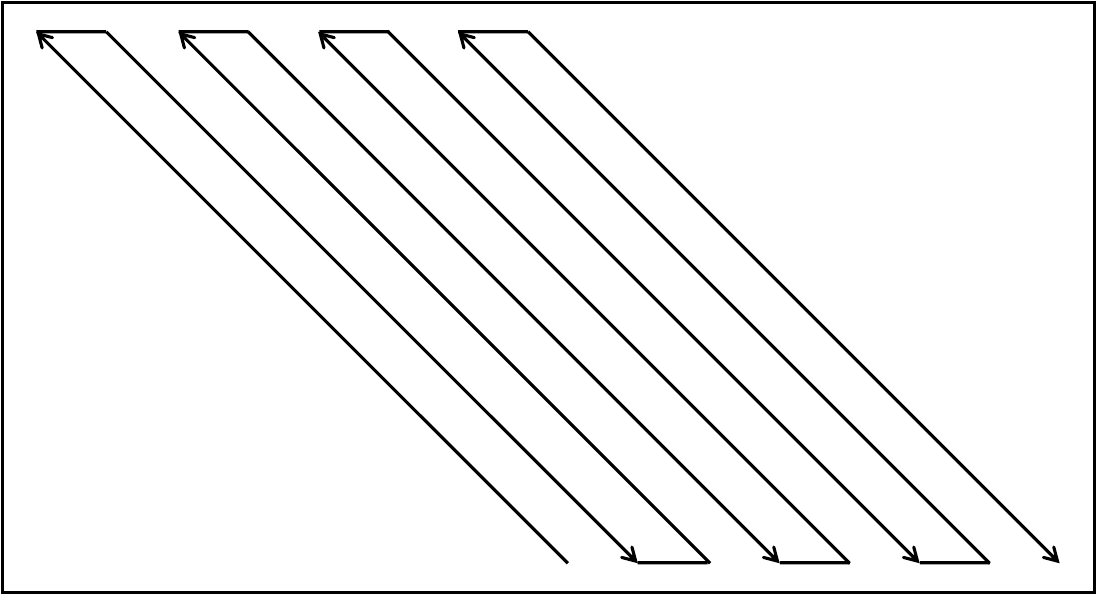}  \\ 
       $(b)$
      \end{minipage} 
	 \vspace{3mm} \\
      \begin{minipage}[t]{0.45\hsize}
        \centering
       \includegraphics[width=4cm]{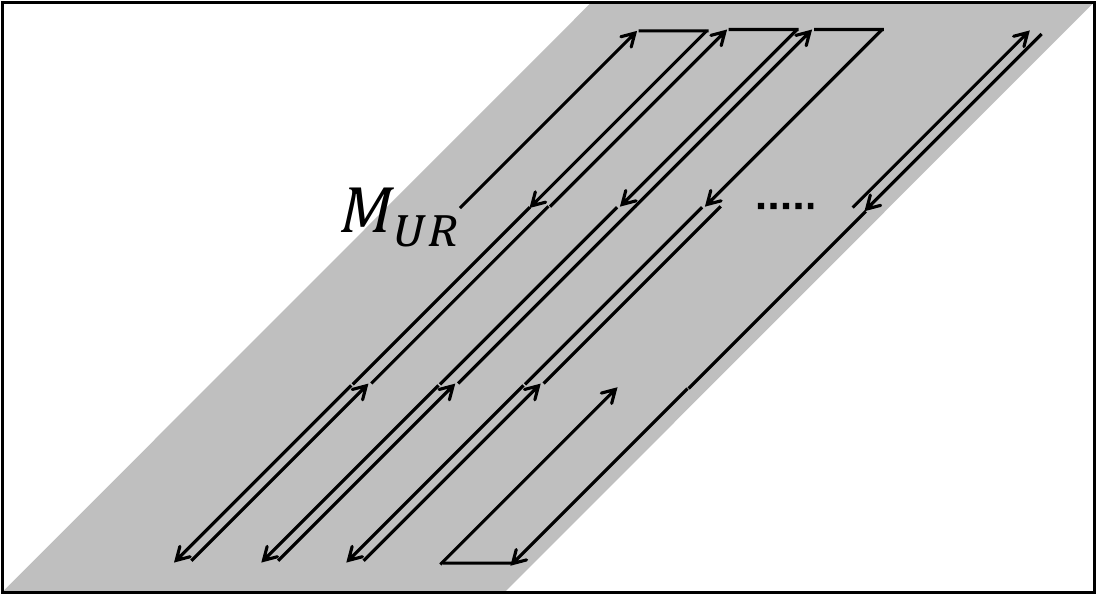}  \\ 
       $(c)$
      \end{minipage} &
      \begin{minipage}[t]{0.45\hsize}
        \centering
       \includegraphics[width=4cm]{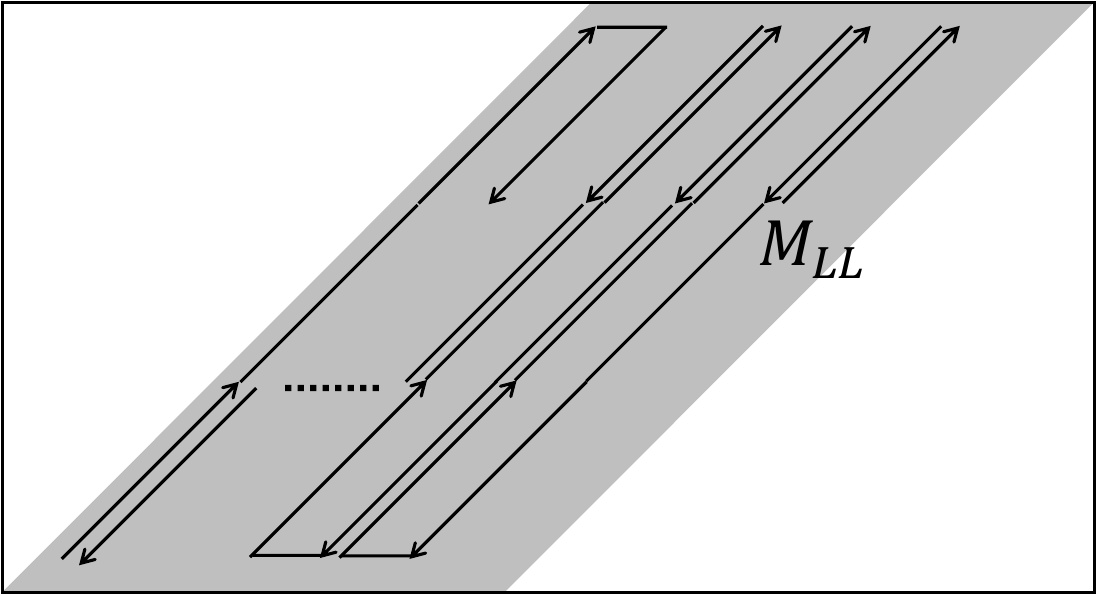}  \\ 
       $(d)$
      \end{minipage} 
    \end{tabular}
\caption{Paths on the field. $(a)$ Path $A$. $(b)$ Path $B$. $(c)$ Forward move on path $A$. $(d)$ Backward move on path $B$.}
 \label{fig:pathAB}
  \end{figure}

In this section, we present a rendezvous algorithm for two MRSs 
each of which consists of five modules with visibility range $7$ and not equipped with 
the global compass. 
The proposed algorithm makes the two MRSs go back and forth 
on a path uniquely fixed in the rectangular field. 
The path starts from the southwest corner and goes at a $\pi/4$ angle to the north wall. 
An MRS follows this path by a zigzag move, which will be presented later. 
Then, it turns on the north wall and goes at a $\pi/4$ angle to the south wall. 
The path is shifted to east during these moves. 
By repeating these moves, the path ends at the northeast corner. 
We call this path ``\emph{path $A$}'' and its mirror image is called ``\emph{path $B$}.'' 
See Figure~\ref{fig:pathAB} $(a)$ and $(b)$.

Due to the lack of the global compass and limited visibility, 
the modules of an MRS cannot recognize which path they are following 
when they move in the middle of the field.  
The difference between path $A$ and path $B$ is the angle formed by the path 
and the north wall (or the south wall). 
When an MRS approaches the north wall along path $A$, 
if we consider the MRS's moving direction is its positive $y$ direction, 
the angle formed by its $x$-axis and the wall is $\pi/4$. 
When an MRS moves along path $B$, 
the angle is $-\pi/4$. 
Based on this fact, the proposed algorithm makes the MRS return to path $A$
when it is not moving on path $A$. 
That is, each module of an MRS does not need the global compass to recognize 
which path it is following. 
We say an MRS is moving with a \emph{correct angle} when its moving direction 
forms angle of $\pi/4$ with the longer wall (i.e., the north or the south wall) lying in front 
of it. 
Otherwise, we say the MRS is moving with a \emph{wrong angle}. 
We call a parallelogram filled by path $A$ the \emph{area of path $A$}. 
We abuse the word ``acute angle'' of the are of path $A$ 
when we refer to the corners of the field that path $A$ touches. 

\begin{figure}[t]
    \begin{tabular}{ccc}
     \multicolumn{3}{c}{
     \begin{minipage}[t]{0.95\hsize}
      \centering
      \includegraphics[width=10cm]{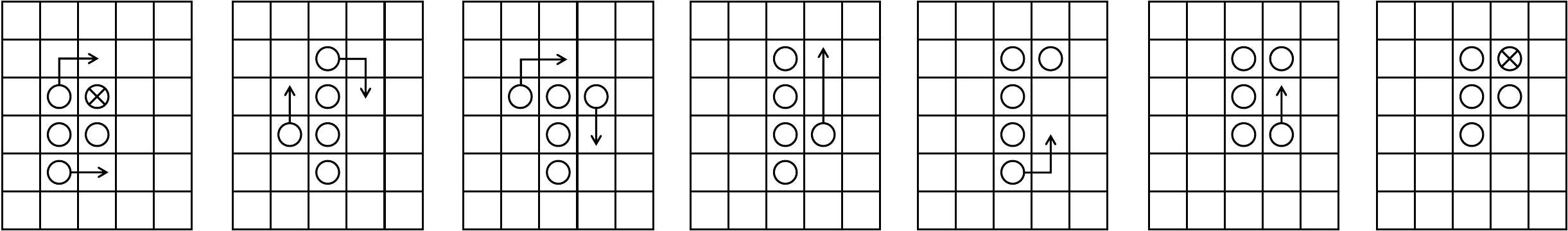}
      \caption{$M_{UR}$. The module with $X$ is a reference point, which shows that the MRS moves to upper right by $M_{UR}$.} 
      \label{fig:M_UR}
     \end{minipage}
     }
     \vspace{3mm} \\ 
     \multicolumn{3}{c}{
     \begin{minipage}[t]{0.95\hsize}
      \centering
      \includegraphics[width=8cm]{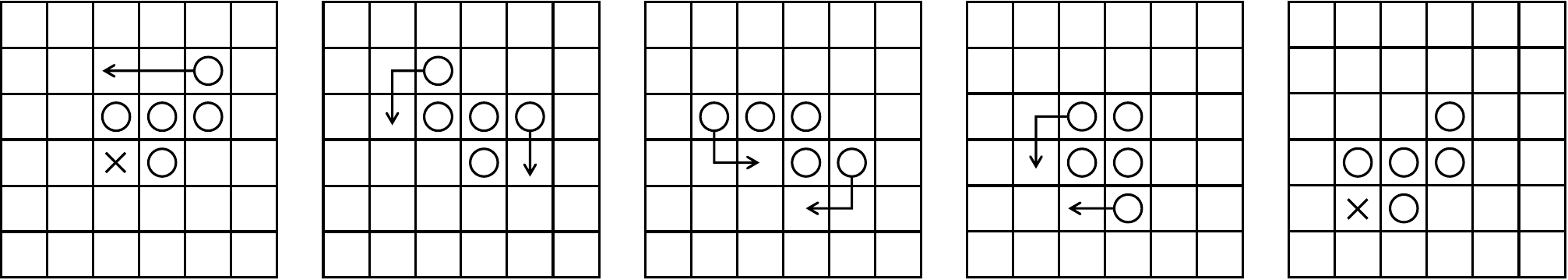}
      \caption{$M_{LL}$. The cell with $X$ is a reference point.} 
      \label{fig:M_LL}
     \end{minipage}
     }
     \vspace{3mm} \\ 
      \begin{minipage}[t]{0.3\hsize}
        \centering
       \includegraphics[width=3cm]{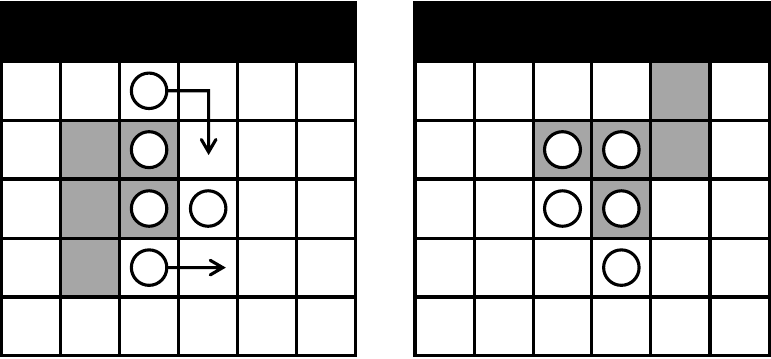}
       \caption{$T_{TW}$} 
       \label{fig:T_TW}
      \end{minipage} &
      \begin{minipage}[t]{0.3\hsize}
        \centering
       \includegraphics[width=2.5cm]{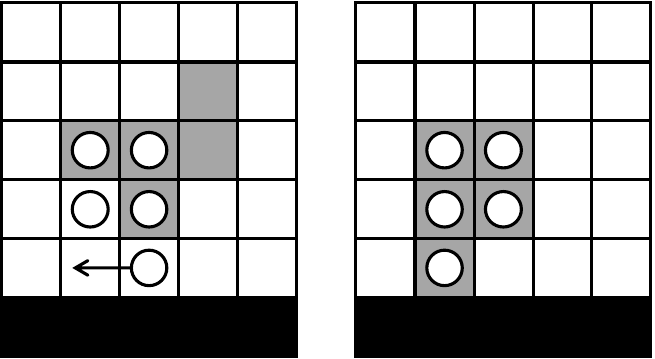}
       \caption{$T_{BW}$} 
       \label{fig:T_BW}
      \end{minipage} &
      \begin{minipage}[t]{0.3\hsize}
        \centering
       \includegraphics[width=3cm]{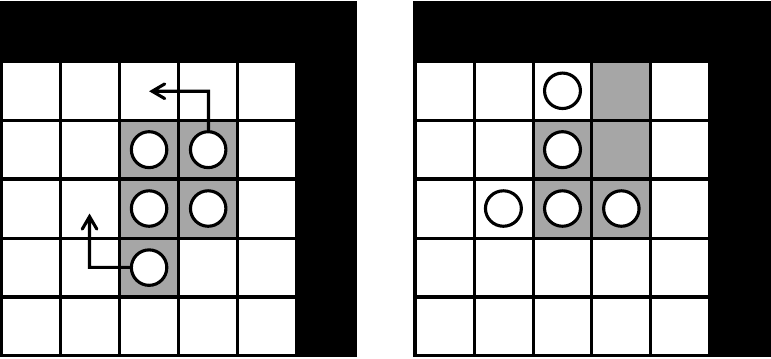}
       \caption{$T_{TC}$} 
       \label{fig:T_TC}
      \end{minipage} 
	     \vspace{3mm} \\
      \begin{minipage}[t]{0.3\hsize}
        \centering
       \includegraphics[width=2.5cm]{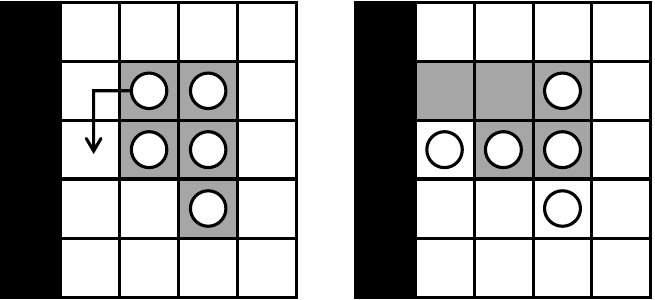}
       \caption{$T_{LW}$} 
       \label{fig:T_LW}
      \end{minipage} &
      \begin{minipage}[t]{0.3\hsize}
       \centering
       \includegraphics[width=2.5cm]{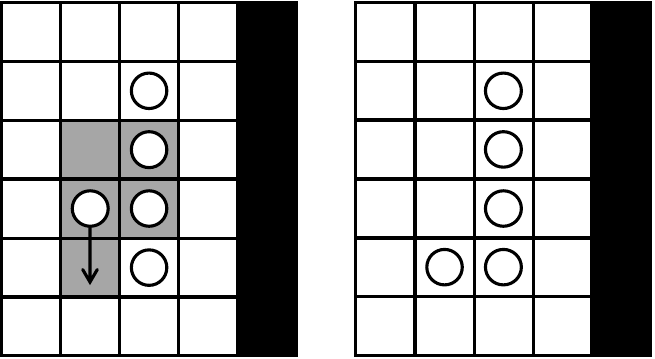}
       \caption{$T_{RW}$} 
       \label{fig:T_RW}
      \end{minipage} &
      \begin{minipage}[t]{0.3\hsize}
        \centering
       \includegraphics[width=4cm]{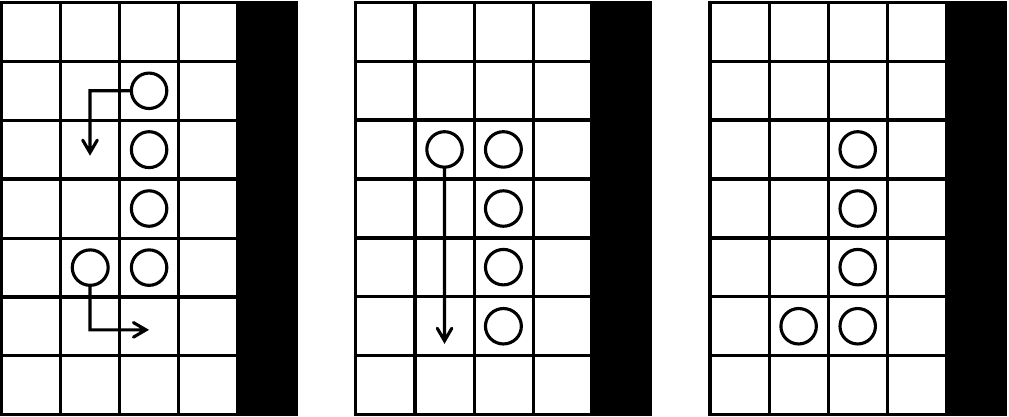}
       \caption{$M_{RW}$} 
       \label{fig:M_RW}
      \end{minipage} 
	     \vspace{3mm} \\ 
     \multicolumn{3}{c}{
     \begin{minipage}[t]{0.95\hsize}
      \centering
      \includegraphics[width=6cm]{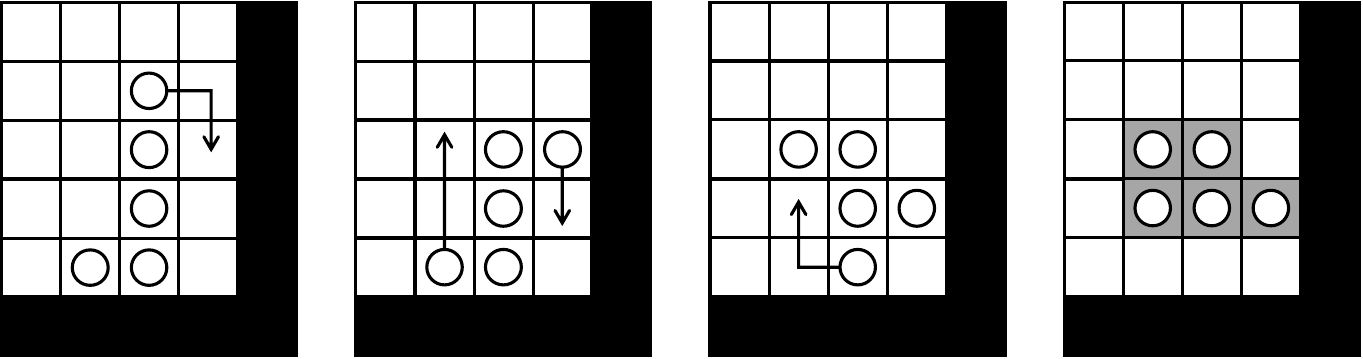}
       \caption{$T_{BC}$} 
       \label{fig:T_BC}
     \end{minipage}
     }
    \end{tabular}
  \end{figure}

We first show the nine unit moves of the proposed algorithm. 
Each of the nine moves consists of a short sequence of states and movements in each state 
as shown in Figure~\ref{fig:M_UR} to \ref{fig:T_BC}. 
The states used in a move are not used in the other moves, 
thus each module can recognize which move the MRS is performing. 
Each module execute the proposed algorithm as follows: 
At each time step, each module checks the state of the MRS to which it belongs, 
identifies the ongoing move, and performs the movement defined for the current state. 
The following is the list of the nine moves.  
\begin{itemize} 
\item Move to upper right denoted by $M_{UR}$ and shown in Figure~\ref{fig:M_UR}. 
The MRS moves diagonally to the right. 
In Figure~\ref{fig:M_UR}, the module with $X$ is a landmark of the MRS. 
After the sequence of $M_{UR}$, the landmark moves up by one cell and right by one cell. 
\item Move to lower left denoted by $M_{LL}$ and shown in Figure~\ref{fig:M_LL}. 
The MRS moves diagonally to the left. 
In Figure~\ref{fig:M_LL}, the cell with $X$ is a landmark of the MRS. 
After the sequence of $M_{LL}$, the landmark moves down by one cell and left by one cell. 
\item Turn on the top wall denoted by $T_{TW}$ and shown in Figure~\ref{fig:T_TW}.  
The MRS approaching to a wall by $M_{UR}$ turns $180$ degrees 
and leaves the wall by $M_{LL}$. 
The gray cells show that the trajectory of the MRS shift by one cell after $T_{TW}$. 
\item Turn on the bottom wall denoted by $T_{BW}$ and shown in Figure~\ref{fig:T_BW}. 
The MRS approaching to a wall by $M_{LL}$ turns $180$ degrees 
and leaves the wall by $M_{UR}$. 
The gray cells show that the trajectory of the MRS shift by one cell after $T_{BW}$. 
\item Turn at a top corner denoted by $T_{TC}$ and shown in Figure~\ref{fig:T_TC}. 
The MRS approaching to a corner by $M_{UR}$ turns $180$ degrees 
and leaves the corner by $M_{UR}$. 
The gray cells show that the trajectory of the MRS does not shift after $T_{TC}$. 
\item Turn on the left wall denoted by $T_{LW}$ and shown in Figure~\ref{fig:T_LW}. 
The MRS approaching a wall with $M_{LL}$ turns $90$ degrees 
and leaves the wall by $M_{UR}$. 
The gray cells show that the trajectory of the MRS shift after $T_{LW}$. 
\item Turn on the right wall denoted by $T_{RW}$ and shown in Figure~\ref{fig:T_RW}. 
The MRS approaching a right wall by $M_{UR}$ turns and starts $T_{RW}$. 
The gray cells show the direction of $M_{UR}$ of the MRS. 
\item  Move down along the right wall denoted by $M_{RW}$ and shown in Figure~\ref{fig:M_RW}. 
The MRS on a right wall moves down along the wall. 
\item Turn at a bottom corner denoted by $T_{BC}$ and shown in Figure~\ref{fig:T_BC}. 
The MRS approaching a corner by $M_{RW}$ turns and 
leaves the corner by $M_{UR}$.
The gray cells show the direction of $M_{UR}$ of the MRS. 
\end{itemize}
As explained above, there are two types of moves, i.e., straight moves and turns. 
A turn changes a straight move to another straight move, 
and Figure~\ref{fig:T_TW} to \ref{fig:T_BC} show the changes by gray cells, 
that show the first state of the straight moves.

These nine moves yield two types of trajectories on path $A$ 
as shown in Figure~\ref{fig:pathAB} $(c)$ and $(d)$. 
Because the modules are not equipped with the global compass, 
an MRS may approach the top wall by executing $M_{LL}$ with a correct angle. 
In this case, the MRS changes its direction by $T_{BW}$ and moves toward the bottom wall 
by $M_{UR}$. 
When the MRS reaches the bottom wall, it changes its direction by $T_{TW}$ 
with sliding its move by one 
and again moves toward the top wall by executing $M_{LL}$. 
This ``backward'' trajectory is shown in Figure~\ref{fig:pathAB} $(d)$. 
When it reaches the bottom corner by executing $M_{UR}$, it turns by $T_{TC}$ 
and moves toward the top wall by executing $M_{UR}$.  
The trajectory is now switched to the ``forward'' trajectory shown in Figure~\ref{fig:pathAB} $(c)$, 
which reaches the top wall by $M_{UR}$. 
It turns on the top wall by $T_{TW}$ with sliding its move by one. 
Then it moves toward the bottom wall by $M_{LL}$. 
When it reaches the bottom wall, it changes its direction by $T_{BW}$ 
and moves toward the top wall by $M_{UR}$. 
When it reaches the top corner, it turns by $T_{TC}$ and it moves toward the 
bottom wall by $M_{UR}$. Now, the trajectory is again switched to the backward trajectory. 
 
\begin{figure}[!htbp]
 \begin{tabular}{cc} 
  \begin{minipage}[t]{0.45\hsize}
   \centering
   \includegraphics[width=5cm]{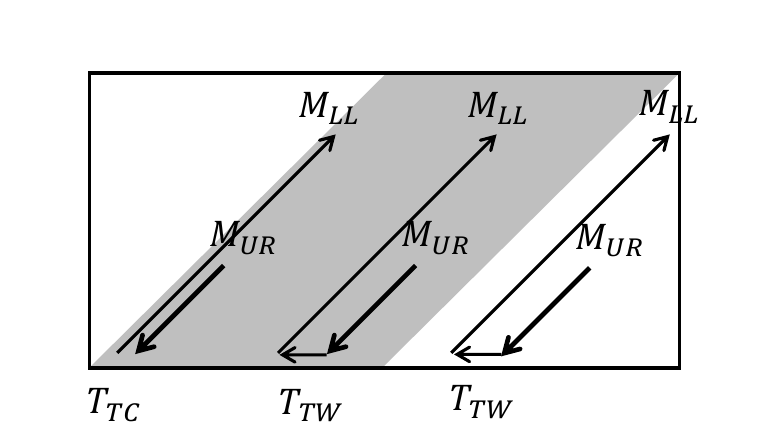} \\  
   $(a)$
  \end{minipage} & 
  \begin{minipage}[t]{0.45\hsize}
   \centering
   \includegraphics[width=5cm]{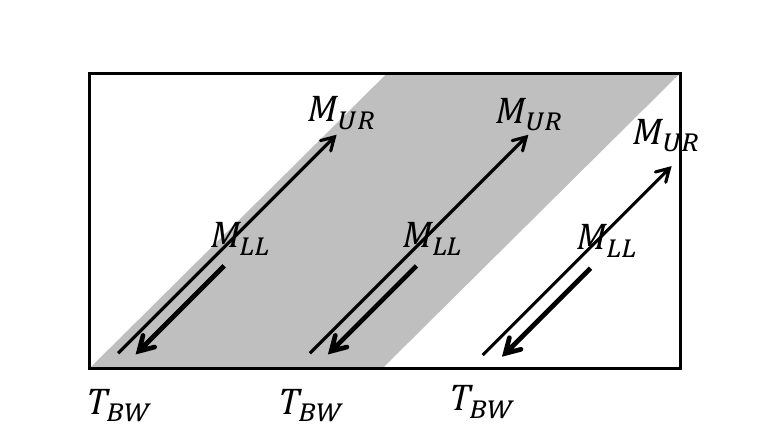} \\
   $(b)$
  \end{minipage} 
  \vspace{3mm} \\
  \begin{minipage}[t]{0.45\hsize}
   \centering
   \includegraphics[width=5cm]{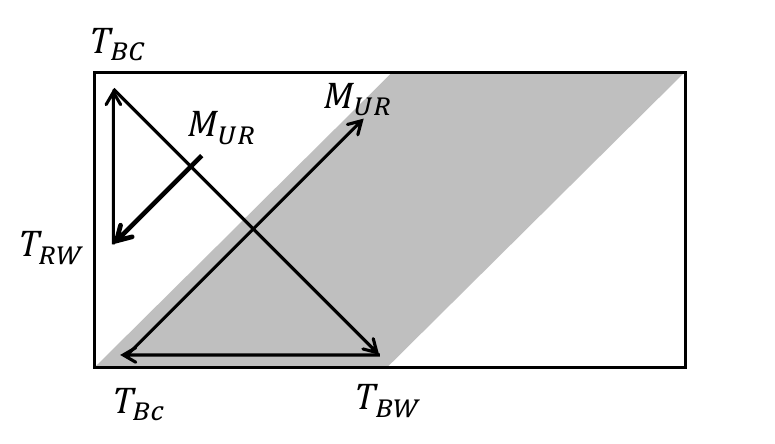} \\
   $(c)$
  \end{minipage} &
  \begin{minipage}[t]{0.45\hsize}
   \centering
   \includegraphics[width=5cm]{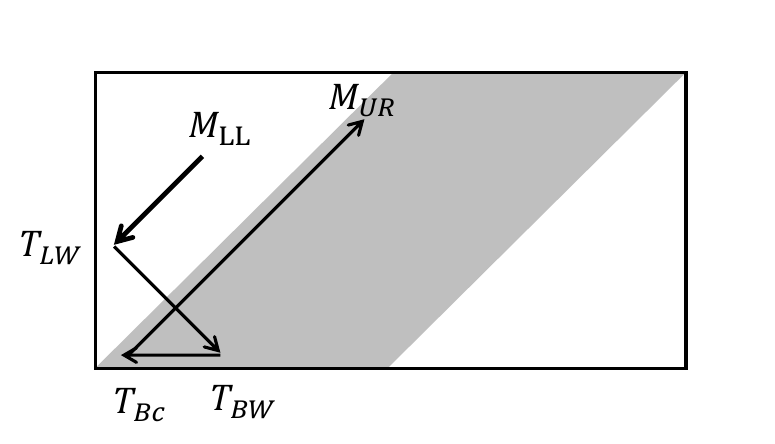} \\
   $(d)$
  \end{minipage}
  \vspace{3mm} \\
  \begin{minipage}[t]{0.45\hsize}
   \centering
   \includegraphics[width=5cm]{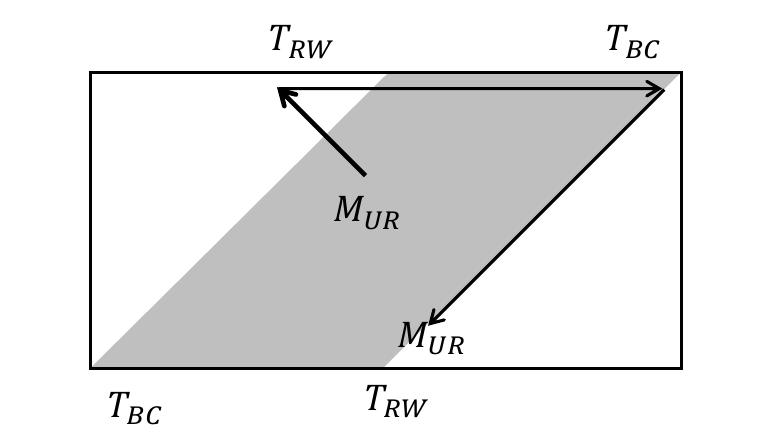} \\
   $(e)$
  \end{minipage} & 
  \begin{minipage}[t]{0.45\hsize}
   \centering
   \includegraphics[width=5cm]{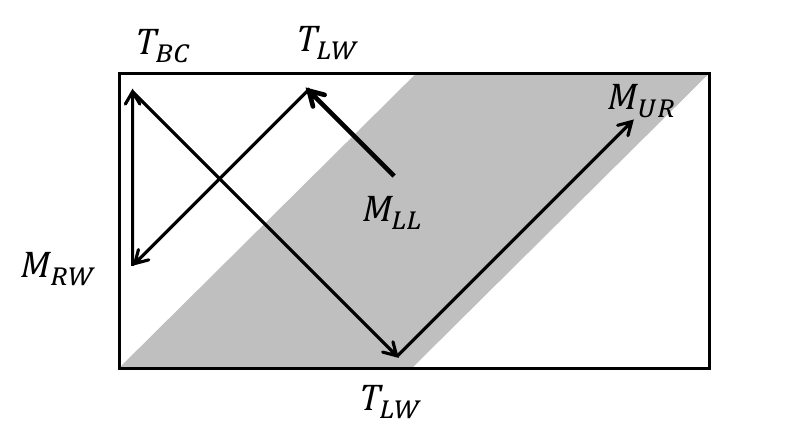} \\
   $(f)$ 
  \end{minipage} 
  \vspace{3mm} \\
  \begin{minipage}[t]{0.45\hsize}
   \centering
   \includegraphics[width=5cm]{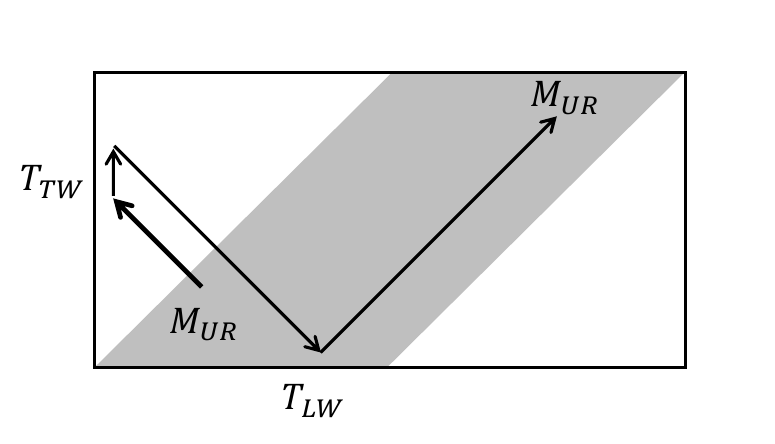} \\
   $(g)$
  \end{minipage} &
  \begin{minipage}[t]{0.45\hsize}
   \centering
   \includegraphics[width=5cm]{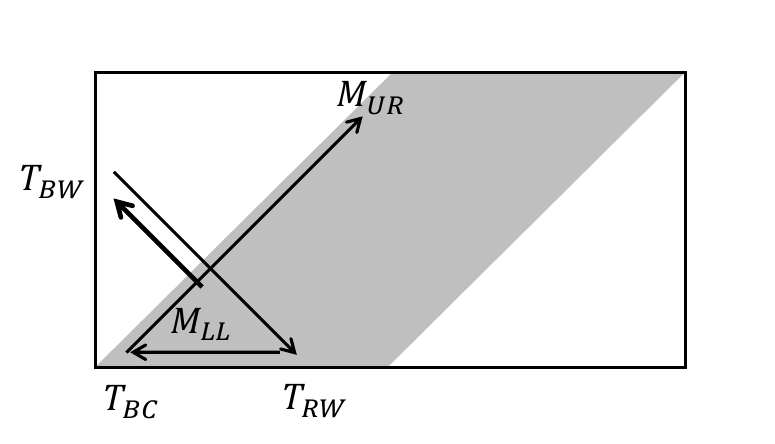} \\
   $(h)$
  \end{minipage} 
  \vspace{3mm} \\
  \begin{minipage}[t]{0.45\hsize}
   \centering
   \includegraphics[width=5cm]{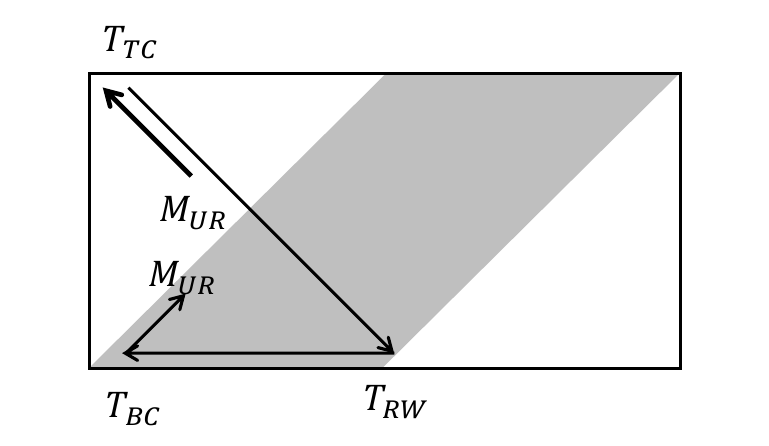} \\
   $(i)$
  \end{minipage} & 
  \begin{minipage}[t]{0.45\hsize}
   \centering
   \includegraphics[width=5cm]{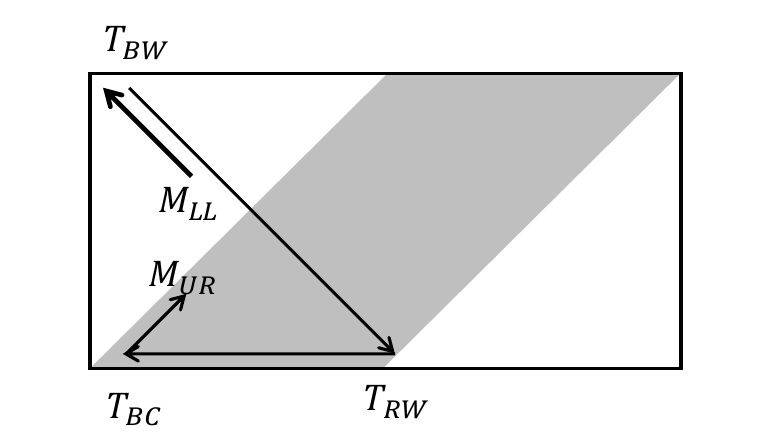} \\ 
   $(j)$
  \end{minipage}
\end{tabular}
\caption{Trajectory of an MRS when it returns to path $A$,}
\label{fig:rendezvou-recover}
\end{figure}

When an MRS is moving with a wrong angle or moving outside of the area of path $A$, 
it can return to path $A$. 
We first show the cases where an MRS moves with a correct angle, but 
it moves outside of the area of path $A$. 
\begin{itemize} 
\item {\bf Case $C1$.} When an MRS approaches a top or bottom wall by $M_{UR}$ 
with a correct angle but outside of the area of path $A$, 
it turns on the wall by $T_{TW}$ and approaches a side wall by $M_{LL}$. 
Hence, this case falls in the following Case $C4$. 
See Figure~\ref{fig:rendezvou-recover} $(a)$. 
\item {\bf Case $C2$.} When an MRS approaches a top or bottom wall by $M_{LL}$ 
with a correct angle but outside of the area of path $A$, 
it turns on the wall by $T_{BW}$ and approaches a side wall by $M_{UR}$. 
Hence, this case falls in the following Case $C3$. 
See Figure~\ref{fig:rendezvou-recover} $(b)$.
\item {\bf Case $C3$.} When the MRS approaches a side wall by $M_{UR}$ 
with a correct angle but outside of the area of path $A$, 
it forms an angle of $-\pi/4$ with the wall. 
Thus, it turns by $T_{RW}$ on the wall and moves along the wall by $M_{RW}$ 
until it reaches a corner. 
Then, it turns by $T_{BC}$ and starts $M_{UR}$. 
This corner is not a corner of path $A$, thus the MRS is still not on path $A$. 
The MRS will eventually reaches a top or bottom wall 
with forming an angle of $-\pi/4$. 
Thus, it performs $T_{RW}$ and moves along the wall by $M_{RW}$ until it reaches a corner. 
Then, it turns by $T_{BC}$ and starts $M_{UR}$ with a correct angle. 
The corner is an acute angle of path $A$, thus the MRS is now moves along path $A$. 
See Figure~\ref{fig:rendezvou-recover} $(c)$.
\item {\bf Case $C4$.} When the MRS approaches a side wall by $M_{LL}$ 
with a correct angle but outside of the area of path $A$, 
it forms an angle of $-\pi/4$ with the wall. 
Thus, it turns by $\pi/4$ by $T_{LW}$ and starts $M_{UR}$. 
The MRS will reach a top or a bottom wall with forming an angle of $-\pi/4$. 
Thus, it turns by $T_{RW}$ and moves along the wall by $M_{RW}$ until it reaches a corner. 
Then, it turns by $T_{BC}$ and starts $M_{UR}$ with a correct angle. 
See Figure~\ref{fig:rendezvou-recover} $(d)$.
\end{itemize}

We then show the cases where an MRS moves with a wrong angle. 
\begin{itemize}
\item {\bf Case $W1$.} When an MRS approaches a top or bottom wall by $M_{UR}$ 
with a wrong angle, 
it forms an angle of $-\pi/4$ with the wall. 
It turns by $T_{RW}$ on the wall and moves along the wall by $M_{RW}$ 
until it reaches a corner. 
Then, it turns by $T_{BC}$ and starts $M_{UR}$ with a right angle. 
The corner is an acute corner of path $A$, thus the MRS now moves along path $A$. 
See Figure~\ref{fig:rendezvou-recover} $(e)$. 
\item {\bf Case $W2$.} When an MRS approaches a top or bottom wall by $M_{LL}$ 
with a wrong angle, 
it forms an angle of $-\pi/4$ with the wall. 
It turns by $\pi/4$ by $T_{LW}$ and starts $M_{UR}$ with the correct angle.
See Figure~\ref{fig:rendezvou-recover} $(f)$. 
\item {\bf Case $W3$.} When an MRS approaches a side wall by $M_{UR}$ 
with a wrong angle, 
it forms an angle of $\pi/4$ with the wall. 
It performs $T_{TW}$ and starts $M_{LL}$ toward the top or bottom wall 
with a wrong angle. 
Thus it falls in Case $W2$. 
See Figure~\ref{fig:rendezvou-recover} $(g)$.
\item {\bf Case $W4$.} When an MRS approaches a side wall by $M_{LL}$ 
with a wrong angle, 
it forms an angle of $\pi/4$ with the wall. 
It performs $T_{BW}$ and starts $M_{UR}$ toward the top or bottom wall 
with a wrong angle. 
Thus, it falls in Case $W1$.
See Figure~\ref{fig:rendezvou-recover} $(h)$. 
\item {\bf Case $W5$.} When an MRS approaches a corner by $M_{UR}$ 
with a wrong angle, 
it performs $T_{TC}$ and moves by $M_{UR}$ toward the top or bottom wall 
with a wrong angle. 
Thus, it falls in Case $W1$. 
See Figure~\ref{fig:rendezvou-recover} $(i)$.
\item {\bf Case $W6$.} When an MRS approaches a corner by $M_{LL}$ 
with a wrong angle, 
it performs $T_{BW}$ and moves by $M_{UR}$ toward the top or bottom wall 
with a wrong angle. 
Thus, it falls in Case $W1$. 
See Figure~\ref{fig:rendezvou-recover} $(j)$.
\end{itemize}

\begin{figure}[t]
  \begin{tabular}{cc}
   \multicolumn{2}{c}{
   \begin{minipage}[t]{0.9\hsize}
    \centering
    \includegraphics[width=10cm]{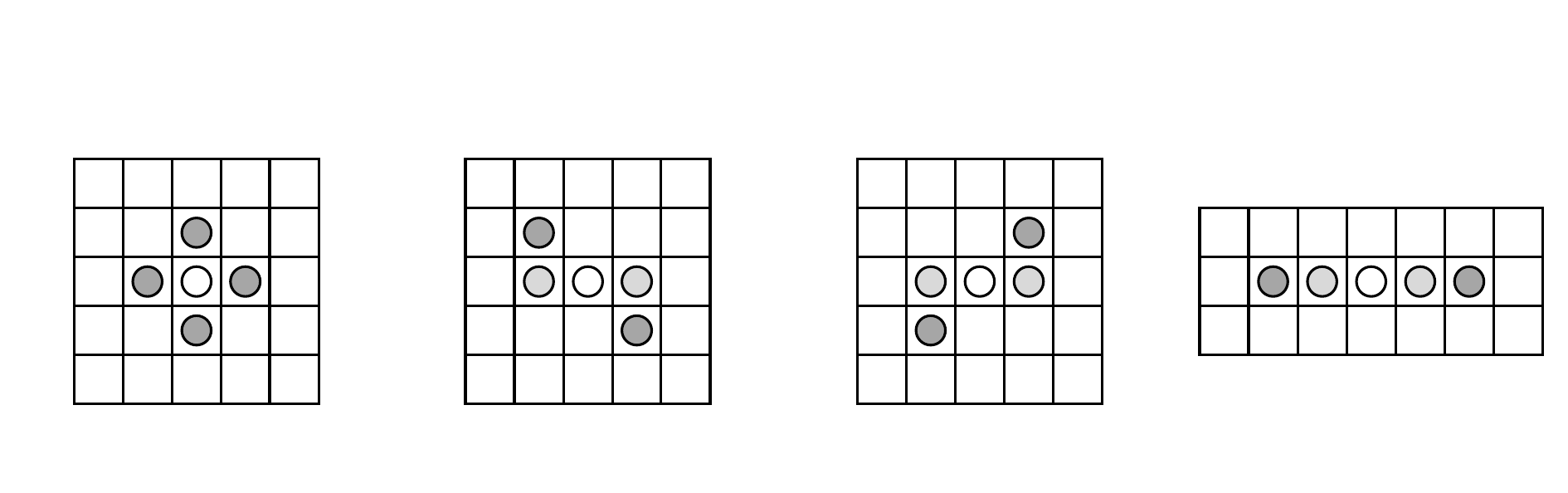}
    \caption{Symmetric states. In each state, a pair of modules that may have a same observation is painted with a same color.} 
    \label{fig:rendezvous-avoid}
   \end{minipage} } 
   \vspace{3mm} \\  
   \begin{minipage}[t]{0.7\hsize}
    \centering
    \includegraphics[width=8cm]{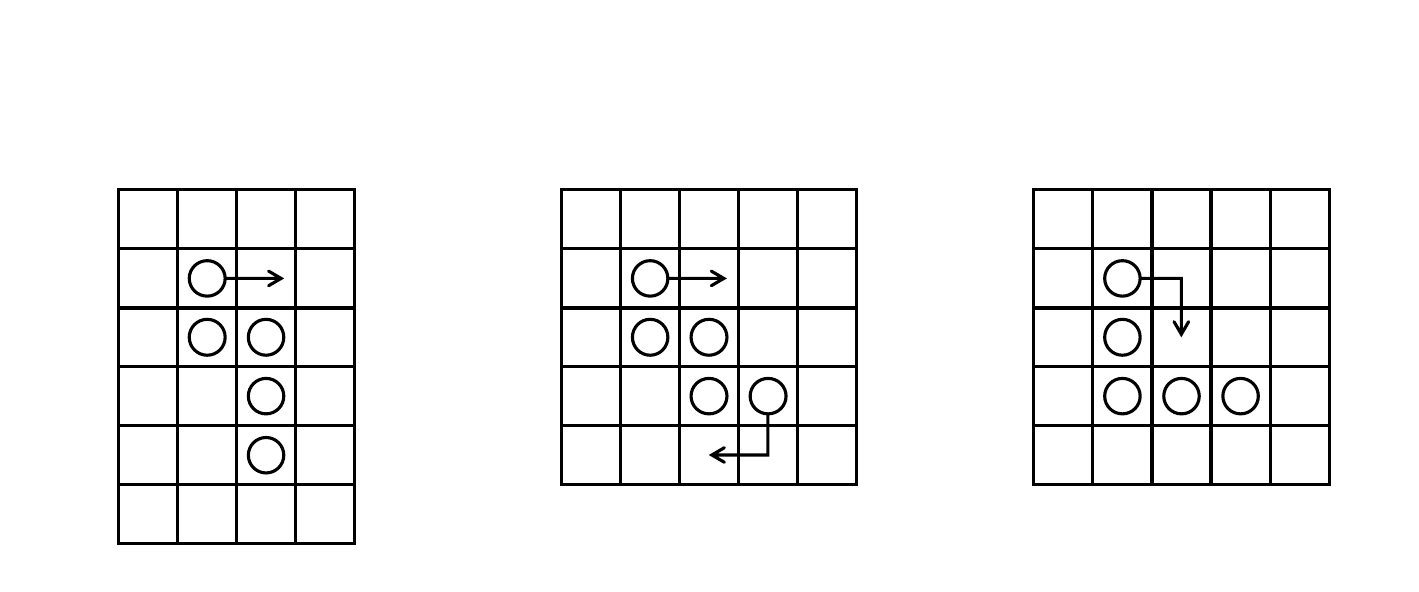} 
    \caption{Exceptional states and transformation to $M_{UR}$} 
    \label{fig:rendezvous-ex} 
   \end{minipage}
   \begin{minipage}[t]{0.25\hsize}
    \centering
    \includegraphics[width=2cm]{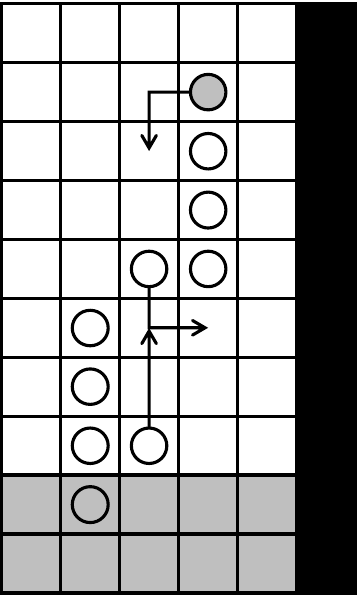} 
    \caption{Collision with visibility range $k=6$} 
    \label{fig:rendezvous_k6} 
   \end{minipage}
\end{tabular}
\end{figure}

As shown in Section~\ref{sec:merge}, five modules not equipped with the 
global compass have $18$ states. 
The proposed rendezvous algorithm does not use the states shown in 
Figure~\ref{fig:rendezvous-avoid} and Figure~\ref{fig:rendezvous-ex}. 
In each state in Figure~\ref{fig:rendezvous-avoid}, 
there exists a pairs of modules that are placed at symmetric positions. 
Thus, the two modules have the same observation in the worst case, and 
move with keeping their symmetry for ever. 
The proposed algorithm does not accept these four symmetric states as an initial state 
and it does not have these states during execution.
Figure~\ref{fig:rendezvous-ex} show the remaining three states 
together with the movements that translate the states into the first state of $M_{UR}$.

In the proposed algorithm, we assumed visibility range of $7$. 
Figure~\ref{fig:rendezvous_k6} shows a collision when the visibility range is $6$. 
In the figure, the MRS in the top, say $R_1$, is performing $M_{RW}$ 
while the MRS in the bottom, say $R_2$, is performing $M_{UR}$. 
The gray module cannot observe the gray cells when its visibility range is $6$, 
hence it cannot determine whether $R_2$ is performing $M_{UR}$ 
or the first exceptional move in Figure~\ref{fig:rendezvous-ex}. 
If $R_2$ is performing $M_{UR}$, $R_1$ and $R_2$ will have a collision 
as shown in Figure~\ref{fig:rendezvous_k6}. 
To avoid this case, we allow the modules to have visibility range of $7$. 

Consequently, we have the following theorem. 

\begin{theorem} 
Two MRSs each of which consists of five modules not equipped with the global compass 
can solve the rendezvous problem 
from an arbitrary initial configuration, where the state of each MRS 
is not symmetric in a field of size $w \times h$ ($w \neq h$).  
\end{theorem}

\section{Merge algorithm} 
\label{sec:merge}

In this section, we present a merge algorithm for an initial configuration, 
where all modules of the two MRSs are in a square area of size $8 \times 8$. 
The initial configurations includes all the terminal configurations of the 
rendezvous algorithm in Section~\ref{sec:rendezvous}. 
In the proposed merge algorithm, we adopt visibility range of $9$. 
Other assumptions on the MRSs are the same as Section~\ref{sec:rendezvous}, i.e., 
the size of each MRS is five and the modules are not equipped with the global compass.

\begin{figure}[!htbp]
    \begin{tabular}{cccc}
      \begin{minipage}[t]{0.2\hsize}
        \centering
       \includegraphics[width=2.5cm]{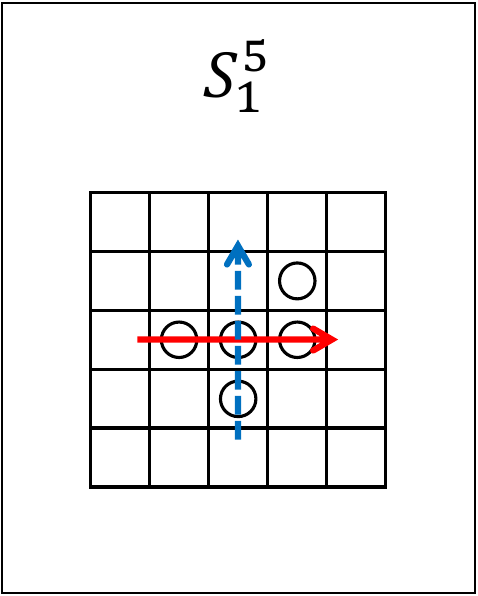}
      \end{minipage} &
      \begin{minipage}[t]{0.2\hsize}
        \centering
       \includegraphics[width=2.5cm]{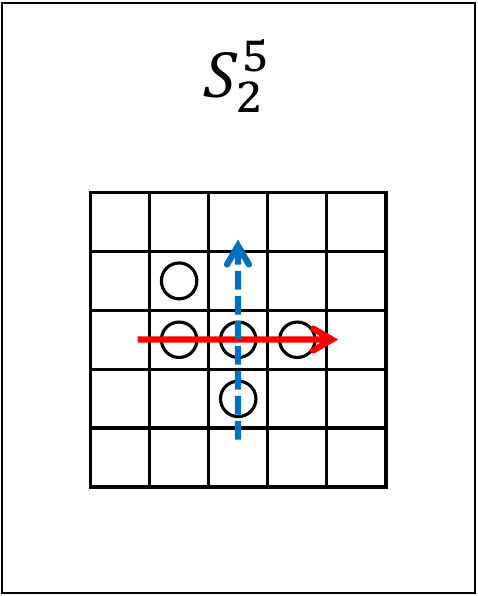}
      \end{minipage} &
      \begin{minipage}[t]{0.2\hsize}
        \centering
       \includegraphics[width=2.5cm]{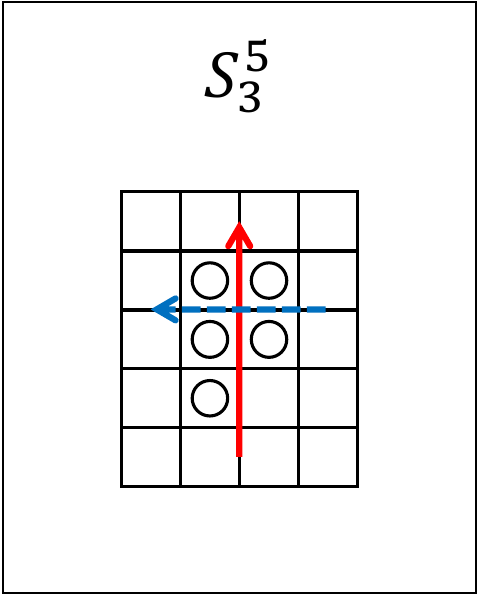}
      \end{minipage} & 
      \begin{minipage}[t]{0.2\hsize}
        \centering
       \includegraphics[width=2.5cm]{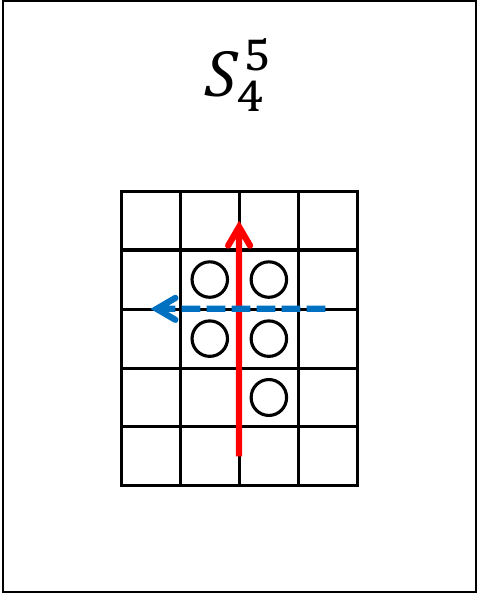}
      \end{minipage} 
     \vspace{3mm} \\
      \begin{minipage}[t]{0.2\hsize}
        \centering
       \includegraphics[width=2.5cm]{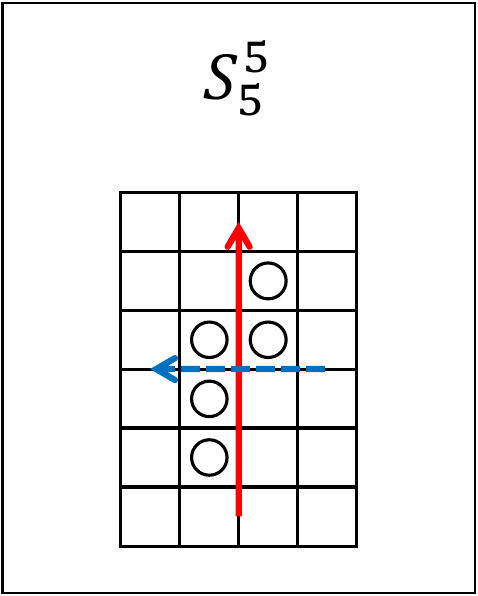}
      \end{minipage} & 
      \begin{minipage}[t]{0.2\hsize}
        \centering
       \includegraphics[width=2.5cm]{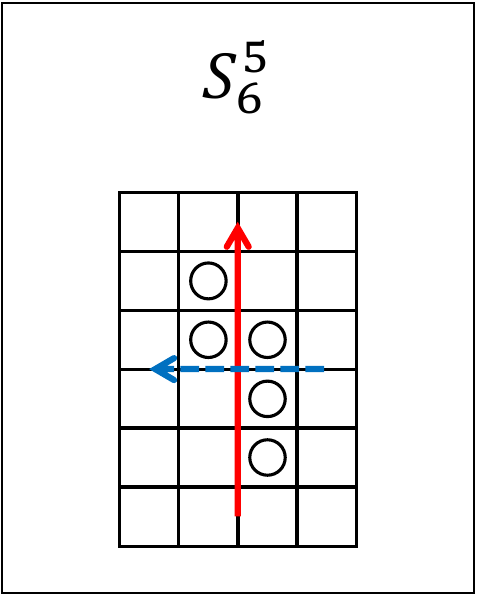}
      \end{minipage} & 
      \begin{minipage}[t]{0.2\hsize}
        \centering
       \includegraphics[width=2.5cm]{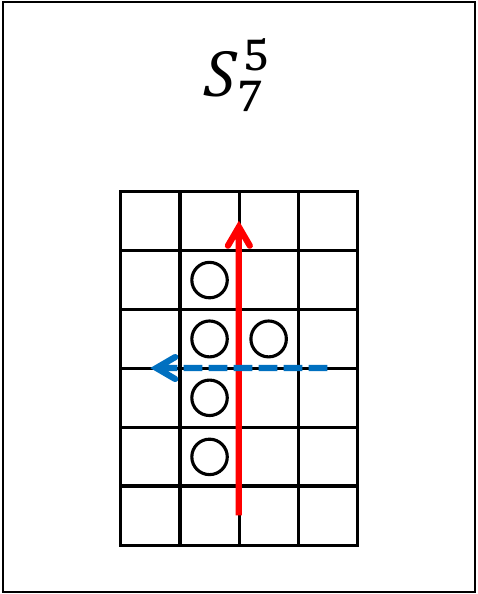}
      \end{minipage} &
      \begin{minipage}[t]{0.2\hsize}
        \centering
       \includegraphics[width=2.5cm]{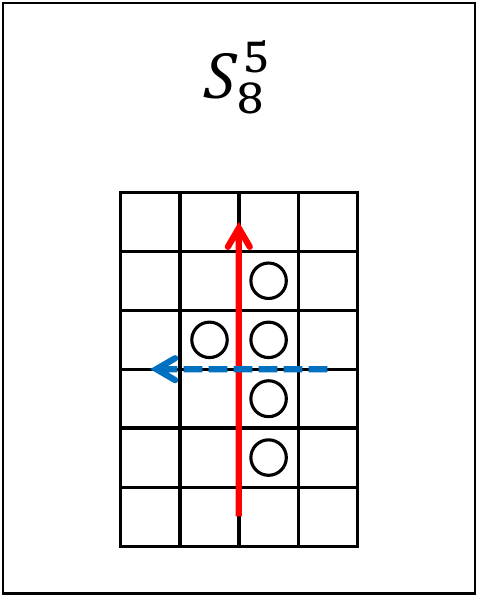}
      \end{minipage} 
     \vspace{3mm} \\
      \begin{minipage}[t]{0.2\hsize}
        \centering
       \includegraphics[width=2.5cm]{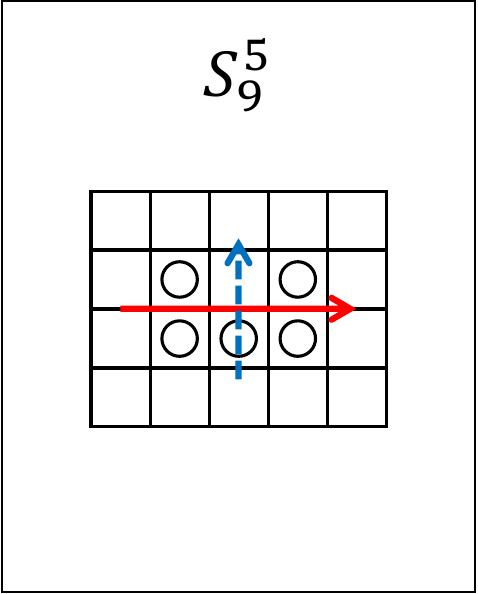}
      \end{minipage} & 
      \begin{minipage}[t]{0.2\hsize}
        \centering
       \includegraphics[width=2.5cm]{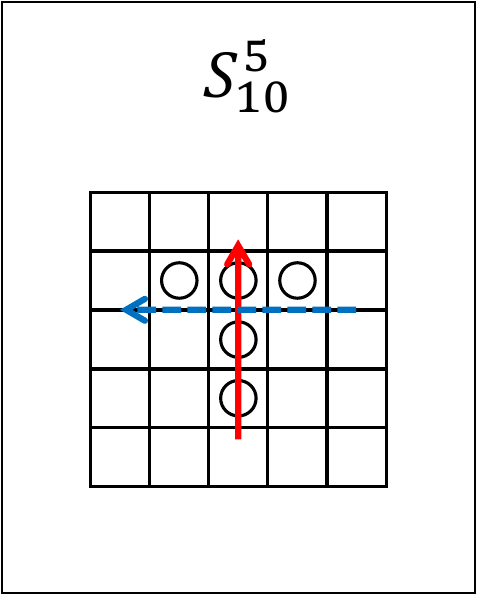}
      \end{minipage} &
      \begin{minipage}[t]{0.2\hsize}
        \centering
       \includegraphics[width=2.5cm]{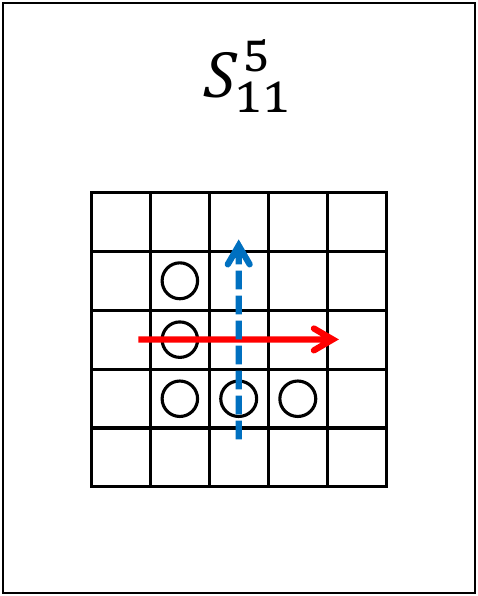}
      \end{minipage} &
      \begin{minipage}[t]{0.2\hsize}
        \centering
       \includegraphics[width=2.5cm]{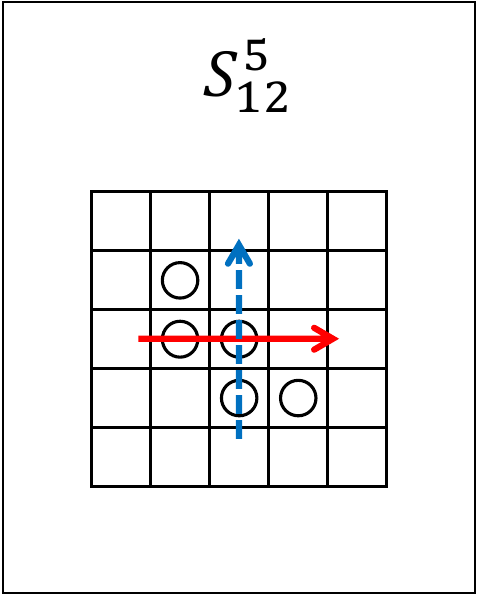}
      \end{minipage} 
     \vspace{3mm} \\
      \begin{minipage}[t]{0.2\hsize}
        \centering
       \includegraphics[width=2.5cm]{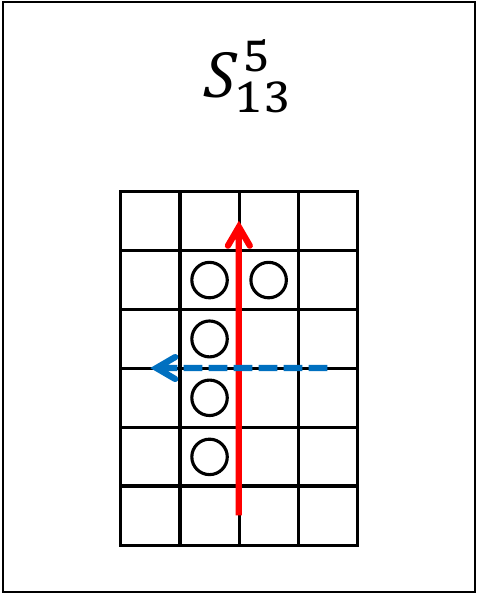}
      \end{minipage} &
      \begin{minipage}[t]{0.2\hsize}
        \centering
       \includegraphics[width=2.5cm]{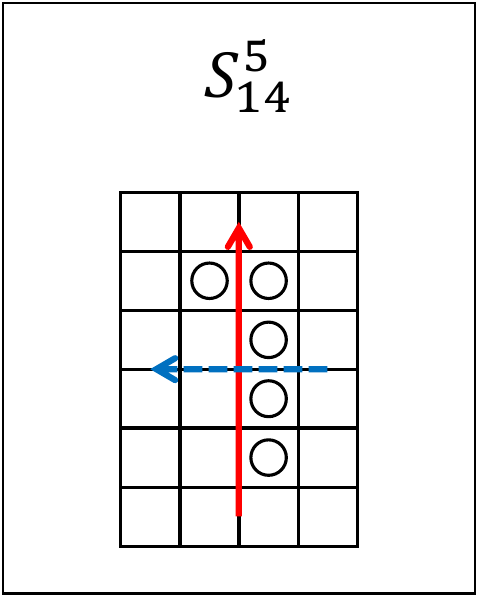}
      \end{minipage} &
      \begin{minipage}[t]{0.2\hsize}
        \centering
       \includegraphics[width=2.5cm]{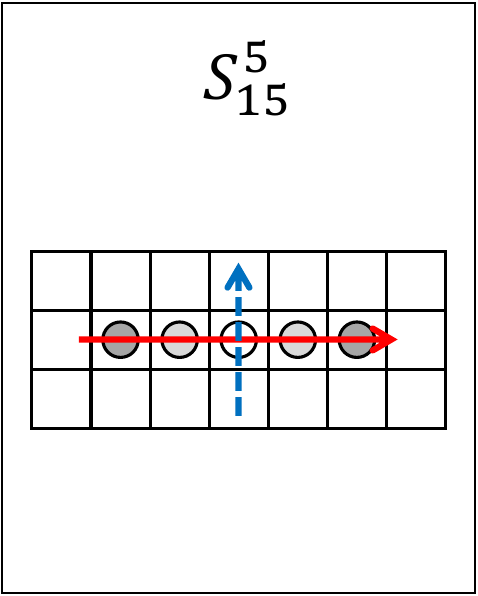}
      \end{minipage} & 
      \begin{minipage}[t]{0.2\hsize}
        \centering
       \includegraphics[width=2.5cm]{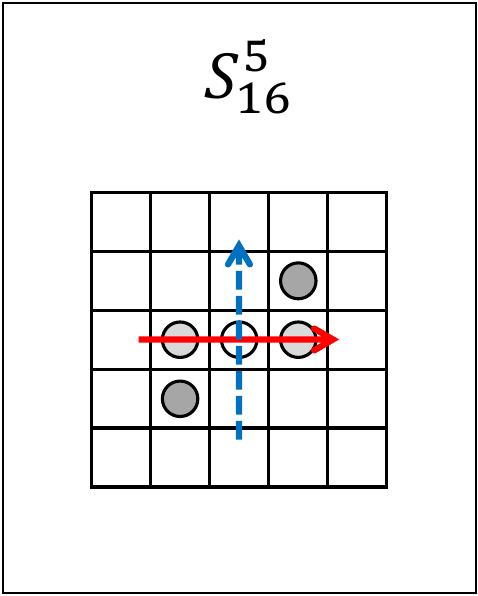}
      \end{minipage} 
     \vspace{3mm} \\
      \begin{minipage}[t]{0.2\hsize}
        \centering
       \includegraphics[width=2.5cm]{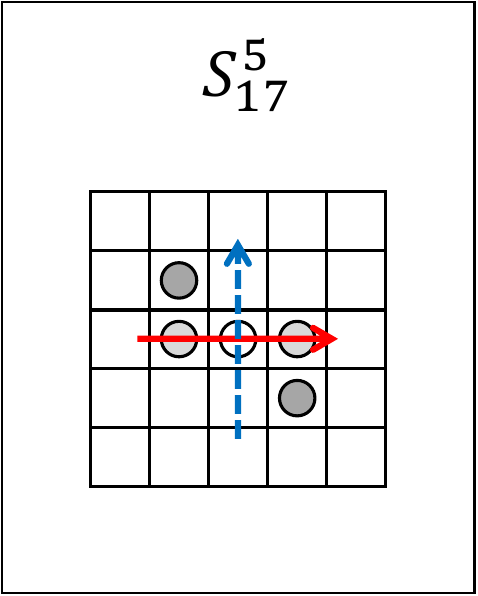}
      \end{minipage} &
      \begin{minipage}[t]{0.2\hsize}
        \centering
       \includegraphics[width=2.5cm]{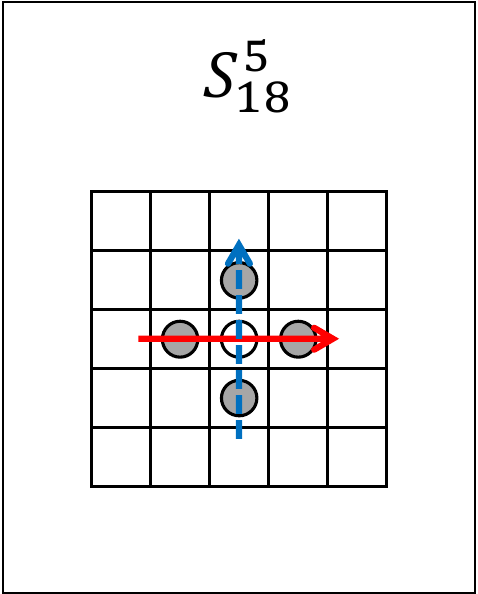}
      \end{minipage} & 
      \begin{minipage}[t]{0.2\hsize}
      \end{minipage} &
      \begin{minipage}[t]{0.2\hsize}
      \end{minipage} 

   \end{tabular}
\caption{$18$ states of an MRS of size five. For each state, its local coordinate system is also shown by the red arrow ($x$-axis) and the blue arrow ($y$-axis).}  
\label{fig:states-5}
  \end{figure}

The main trick of the proposed algorithm is a labeling of all the $18$ states of an MRS 
that consists of five modules not equipped with the global compass. 
Figure~\ref{fig:states-5} shows all these states with their labeling in the form of $S_i^5$. 
We say label $S_i^5$ is smaller than label $S_j^5$ if $i < j$. 
Figure~\ref{fig:states-5} also shows the local coordinate system in each state, 
which is used to break the ties when the two MRSs are in the same state.  

When the initial states of the two MRSs are different, 
the proposed algorithm makes the MRS with the smaller label move 
to the other MRS, that stays in its initial position. 
Otherwise, the proposed algorithm makes the two MRSs to compare their observations 
translated by their local coordinate systems. 
The \emph{view} of an MRS in state $S^5_i$ is a sequence 
$((x_1, y_1), (x_2, y_2), \ldots)$ of coordinates of the center of all modules 
in its visibility 
that satisfies $(i)$ $x_i \leq x_{i+1}$ or $(ii)$ $x_i = x_{i+1}$ and $y_i < y_j$ 
for all $i = 0, 1, 2, \ldots$. 
In states $S^5_{15}, S^5_{16}, S^5_{17}, S^5_{18}$, 
there are multiple choices for the local coordinate systems 
due to the symmetry of the states.  
In this case, we select the local coordinate system that minimizes the view 
in the lexicographic ordering. 
When the views of the two MRSs are different, 
the proposed algorithm makes the MRS with the smaller view 
move to the other MRS, that stays in its initial position. 
Finally, if the two MRSs are in the same state with the same view, 
(i.e., their states and coordinate systems are symmetric), 
they approach each other until they merge int one MRS. 
 
\begin{figure}[ht]
\centering 
\includegraphics[width=12cm]{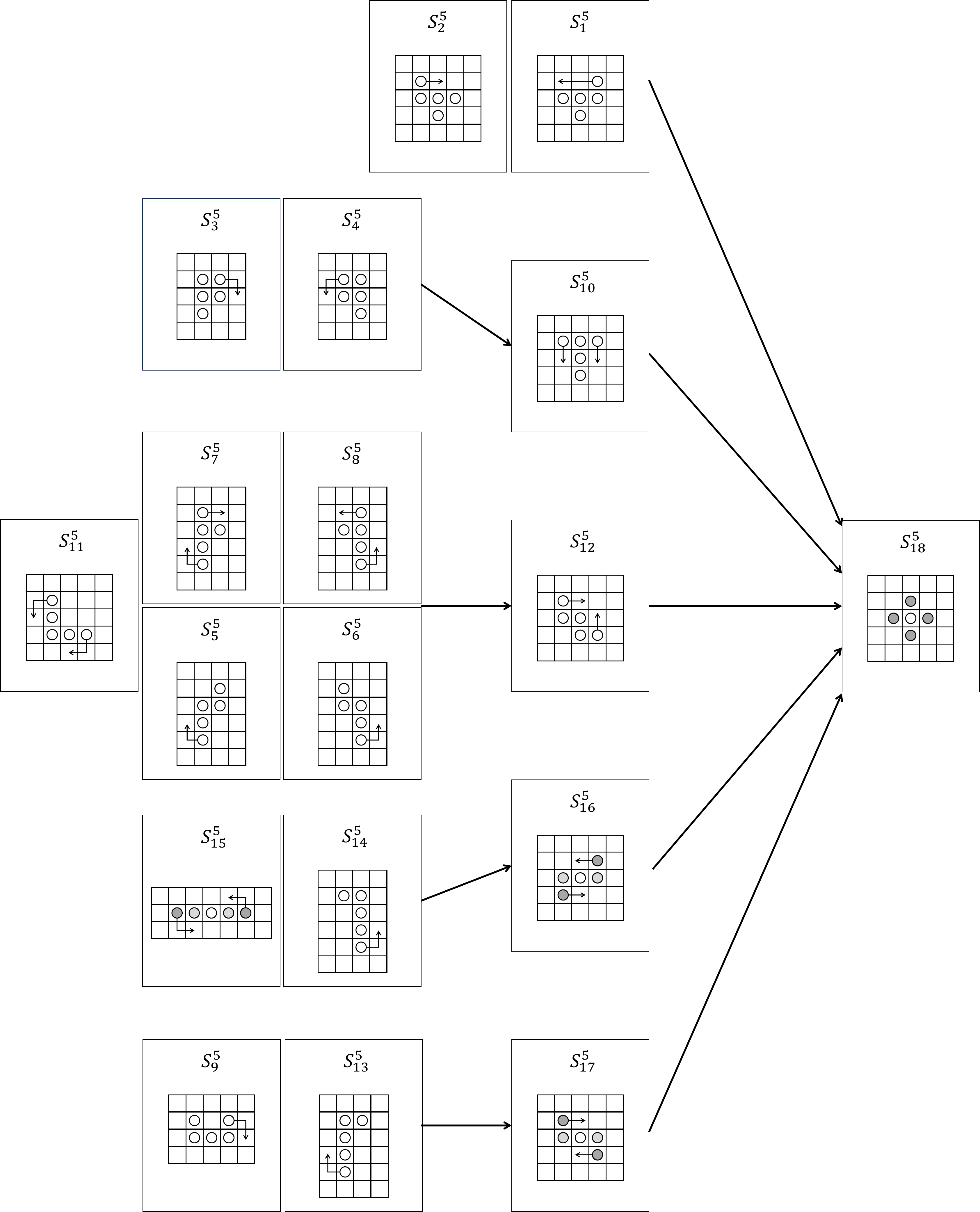}
\caption{Transformation to $S^5_{18}$ with increasing the label of an MRS} 
\label{fig:Inc_TD}
\end{figure}

\begin{figure}[ht]
\centering 
\includegraphics[width=12cm]{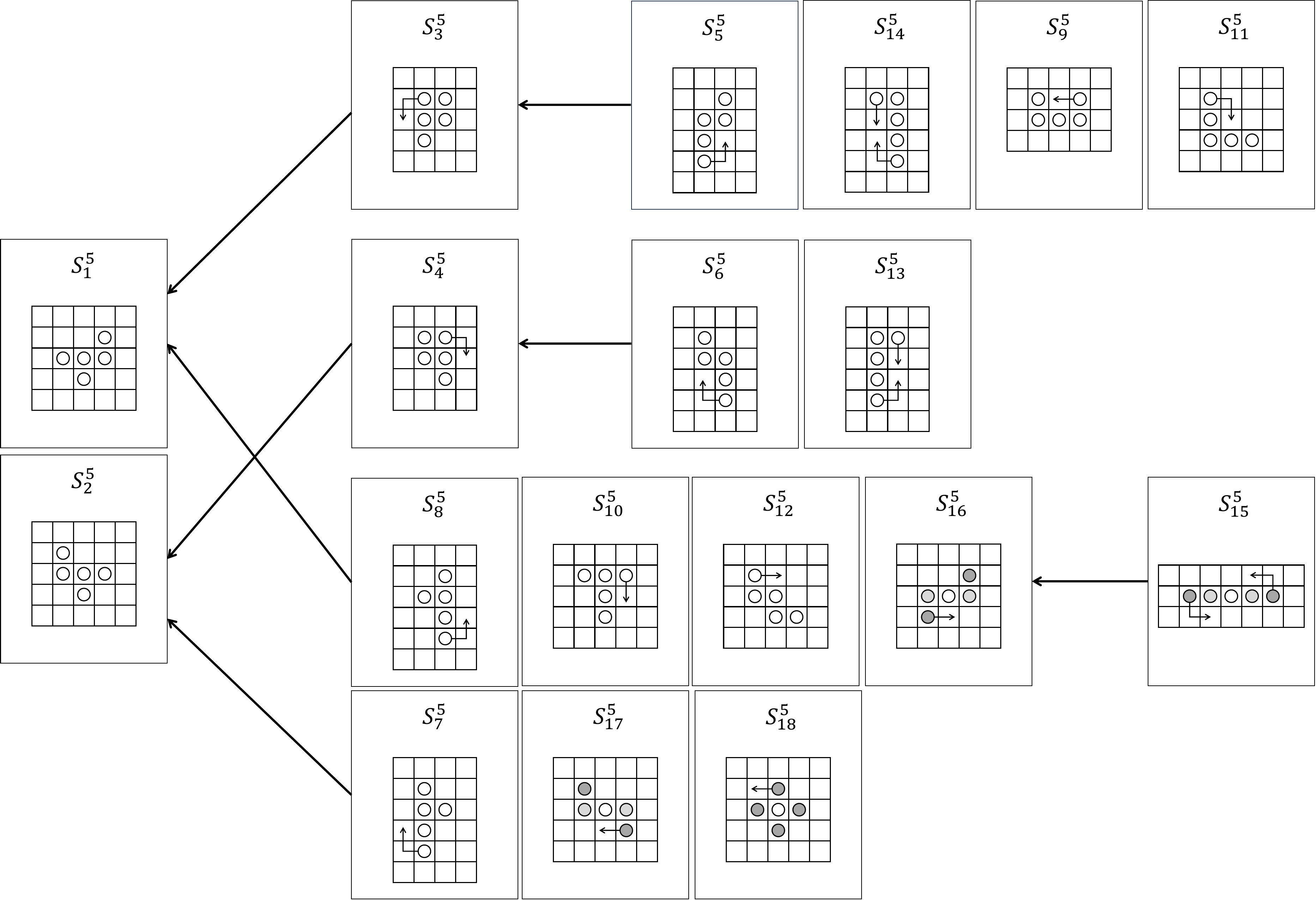}
\caption{Transformation to $S^5_1$ or $S^5_2$ with decreasing the label of an MRS} 
\label{fig:Dec_TD}
\end{figure}

To make the MRSs remember their initial roles, the proposed algorithm 
first makes the MRS with the larger label or larger view 
change its state to $S^5_{18}$ by increasing its label. 
Figure Figure~\ref{fig:Inc_TD} shows the transformations and movements in each state. 
After that, the MRS with the smaller label or smaller view 
changes its state to $S^5_1$ or $S^5_2$ by decreasing its label. 
Figure Figure~\ref{fig:Dec_TD} shows the transformations and movements in each state. 
Then, the MRS approaches to the other MRS by repeating transformations  
between $S^5_1$ and $S^5_2$. 


\begin{figure}[t]
    \begin{tabular}{ccccc}
      \begin{minipage}[t]{0.18\hsize}
        \centering
       \includegraphics[height=2.2cm]{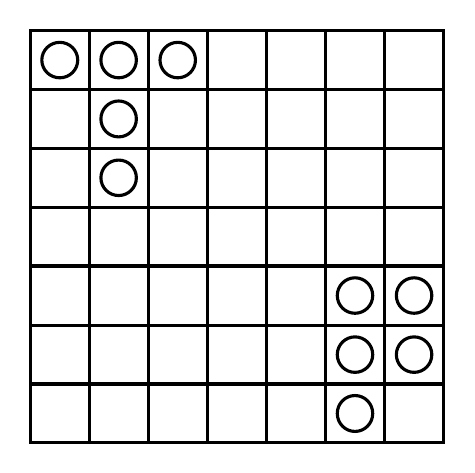} \\
       $(a)$
      \end{minipage} &
      \begin{minipage}[t]{0.18\hsize}
        \centering
       \includegraphics[height=2.2cm]{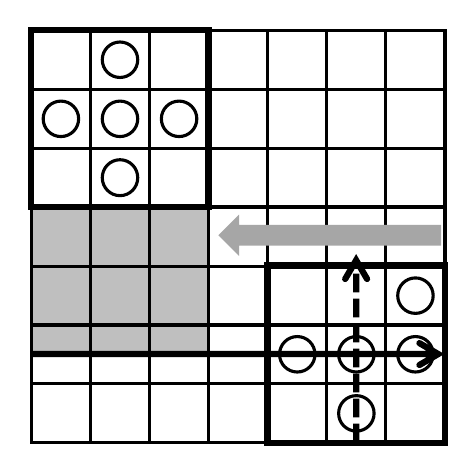} \\
       $(b)$
      \end{minipage} &
      \begin{minipage}[t]{0.18\hsize}
        \centering
       \includegraphics[height=2.2cm]{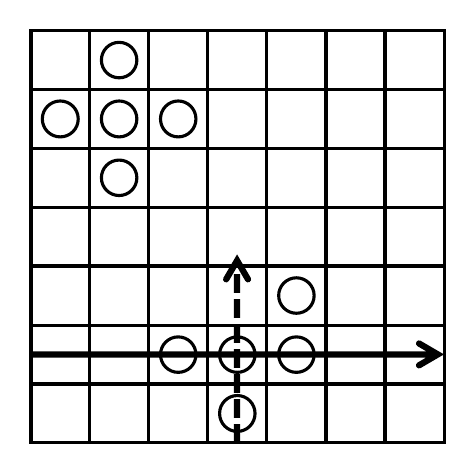} \\
       $(c)$
      \end{minipage} &
      \begin{minipage}[t]{0.18\hsize}
        \centering
       \includegraphics[height=2.2cm]{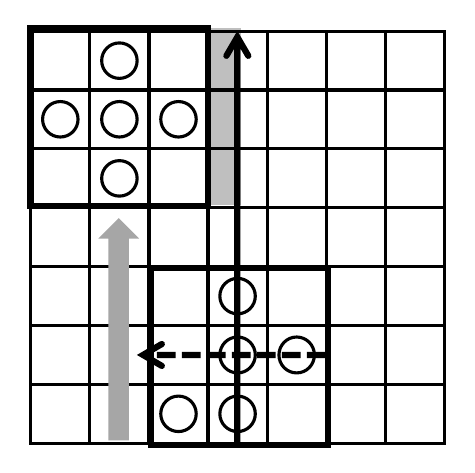} \\
       $(d)$
      \end{minipage} &
      \begin{minipage}[t]{0.18\hsize}
        \centering
       \includegraphics[height=2.2cm]{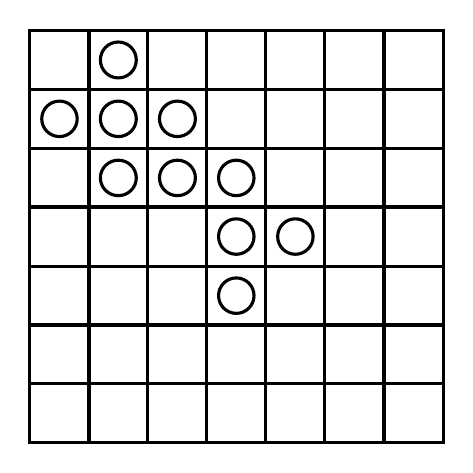} \\
       $(e)$
      \end{minipage} 
    \end{tabular} 
 \caption{Progress of the merge algorithm. $(a)$ An initial configuration. The MRS in $S^5_{10}$ changes its state to $S^5_{18}$. $(b)$ The MRS in $S^5_1$ moves by $M_{x-}$. $(c)$ The MRSs are now $x$-overlapping and the MRS in $S^5_1$ performs $T_{S^5_1}$. $(d)$ The MRS in $S^5_2$ moves by $M_{x+}$. $(e)$ Two MRSs establish connectivity. } 
 \label{fig:merge}
\end{figure}

\begin{figure}[t]
    \begin{tabular}{cc}
      \begin{minipage}[t]{0.45\hsize}
        \centering
       \includegraphics[width=5cm]{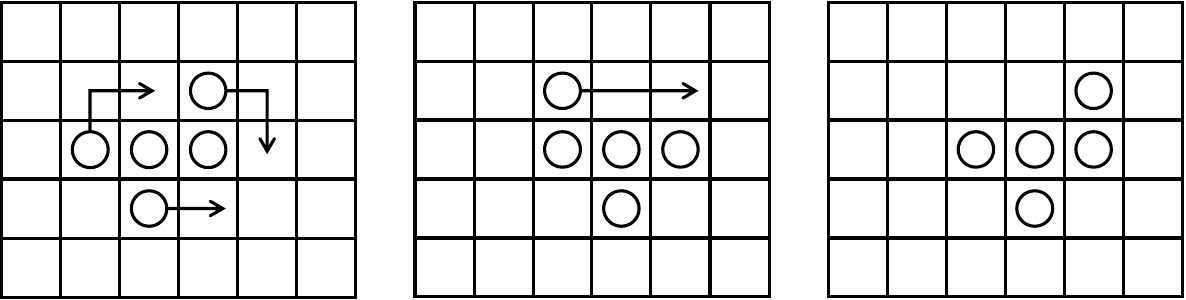}
       \caption{$M_{x+}$} 
       \label{fig:horizontal-p}
      \end{minipage} &
      \begin{minipage}[t]{0.45\hsize}
        \centering
       \includegraphics[width=5cm]{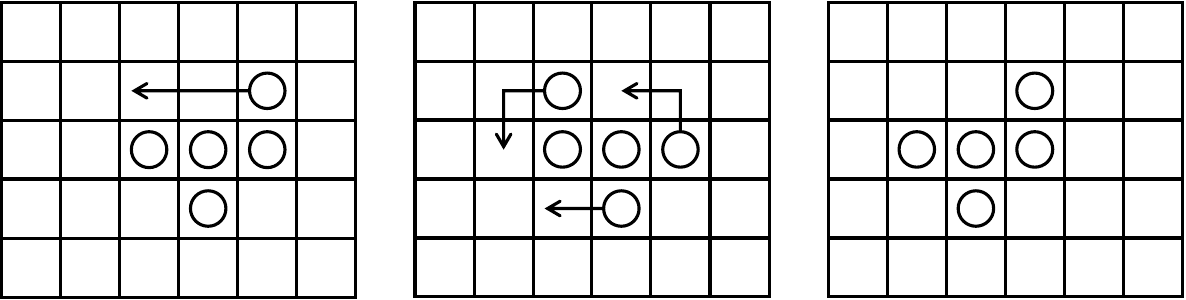}
       \caption{$M_{x-}$}
       \label{fig:horizontal-n} 
      \end{minipage} 
\vspace{3mm} \\ 
      \begin{minipage}[t]{0.45\hsize}
        \centering
       \includegraphics[width=3cm]{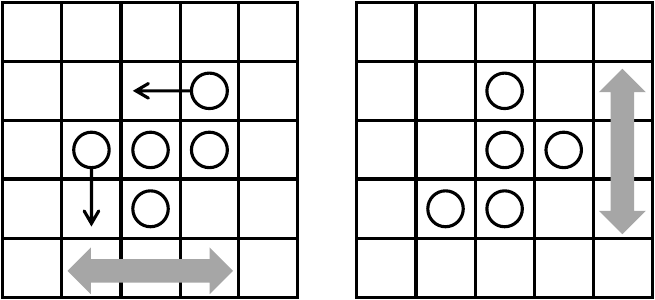} 
       \caption{$T_{S^5_1}$. Gray arrows show the moving direction by $M_{x+}$ and $M_{x-}$. } 
       \label{fig:rotation_s5_1}
      \end{minipage} &
      \begin{minipage}[t]{0.45\hsize}
        \centering
       \includegraphics[width=3cm]{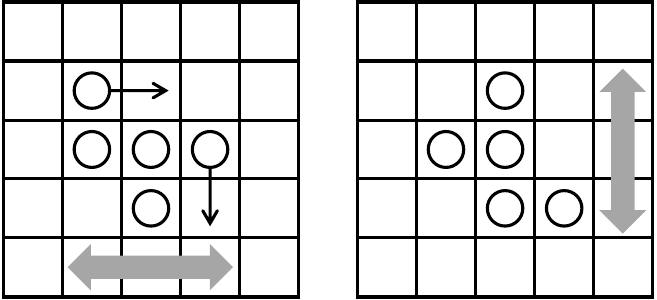} 
       \caption{$T_{S^5_2}$.}
       \label{fig:rotation_s5_2} 
      \end{minipage} 
    \end{tabular}
  \end{figure}

Figure~\ref{fig:merge} shows an execution of the proposed algorithm. 
Consider an initial configuration of two MRSs $R_1$ in $S^5_3$ and $R_2$ in $S^5_{10}$. 
The proposed algorithm first makes $R_2$ change its state to $S^5_{18}$ 
by monotonically increasing its label. 
After that, $R_1$ changes its state to $S^5_1$. 
Then, $R_1$ approaches $R_2$ by horizontal moves 
with respect to its local coordinate system. 
The horizontal movement in the positive direction is denoted by $M_{x+}$ and 
shown in Figure~\ref{fig:horizontal-p} and 
that in the negative direction is denoted by $M_{x-}$ and 
shown in Figure~\ref{fig:horizontal-n}. 
When a module of $R_1$ computes its next move, it considers smallest enclosing rectangles of 
$R_1$ and $R_2$. 
For example, the smallest enclosing rectangle of $S^5_{18}$ is a square of size $3 \times 3$ 
enclosing the five modules.  
Then, it considers the projections of the smallest enclosing rectangles of $R_1$ and $R_2$ 
to the $x$-axis of $R_1$. 
See Figure~\ref{fig:merge} $(b)$ as an example. 
We say $R_1$ is \emph{$x$-overlapping} with $R_2$, 
if there is some overlap of the projections of the smallest enclosing rectangles. 
If $R_1$ is not $x$-overlapping with $R_2$, $R_1$ moves horizontally by $M_{x+}$ or $M_{x-}$ 
until it becomes $x$-overlapping with $R_2$ (Figure~\ref{fig:merge} $(c)$). 
Then, $R_1$ changes its moving direction by a turn (Figure~\ref{fig:merge} $(d)$). 
There are two turns; one is denoted by $T_{S^5_1}$ shown in Figure~\ref{fig:rotation_s5_1} 
and the other is denoted by $T_{S^5_2}$ shown in Figure~\ref{fig:rotation_s5_2} 
After the turn, if $R_2$ is not $x$-overlapping with $R_1$,  
$R_1$ moves horizontally until it merges with $R_2$ (Figure~\ref{fig:merge} $(e)$). 
During the move of $R_1$, no module cause a collision with the modules of $R_2$. 
For example, consider the case where 
the upper right module in $S^5_1$ of $M_{x+}$ cause a collision by its rotation. 
In this case, the colliding module of $R_2$ is already side-adjacent to some module of $R_1$. 
In the same way, we can check that all movements of $M_{x+}$ and $M_{x-}$ does not 
yield any collision.

\begin{figure}[ht]
  \begin{tabular}{cc}
   \begin{minipage}[t]{0.45\hsize}
    \centering
    \includegraphics[height=1.5cm]{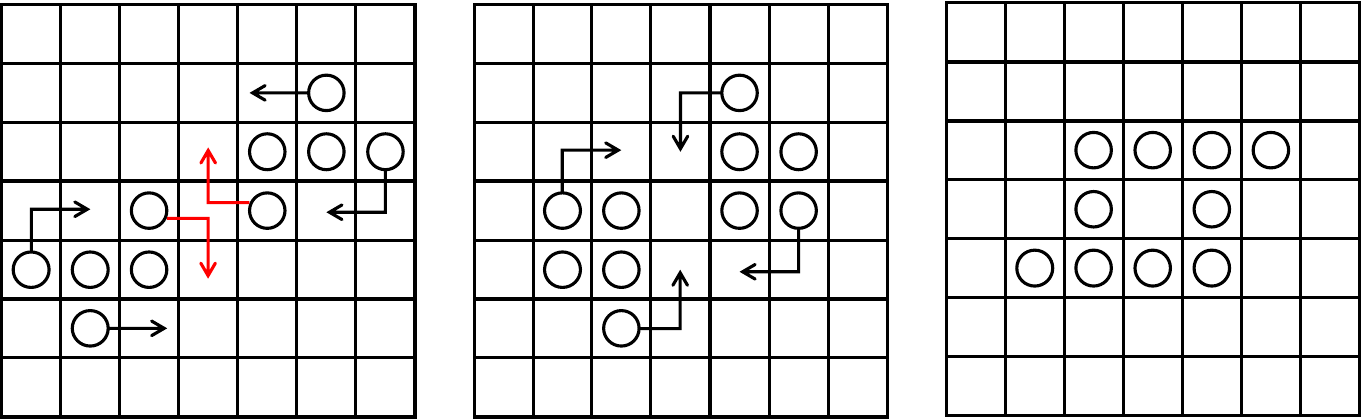} \\
    $(a)$ 
   \end{minipage} &
   \begin{minipage}[t]{0.45\hsize}
    \centering
    \includegraphics[height=1.5cm]{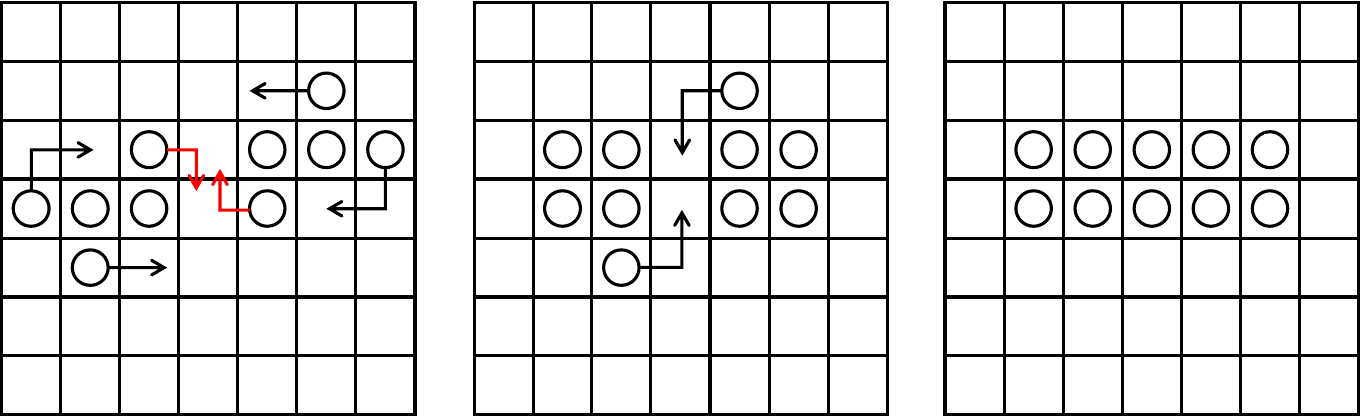} \\
    $(b)$ 
   \end{minipage}
   \vspace{3mm} \\ 
   \begin{minipage}[t]{0.45\hsize}
    \centering
    \includegraphics[height=1.5cm]{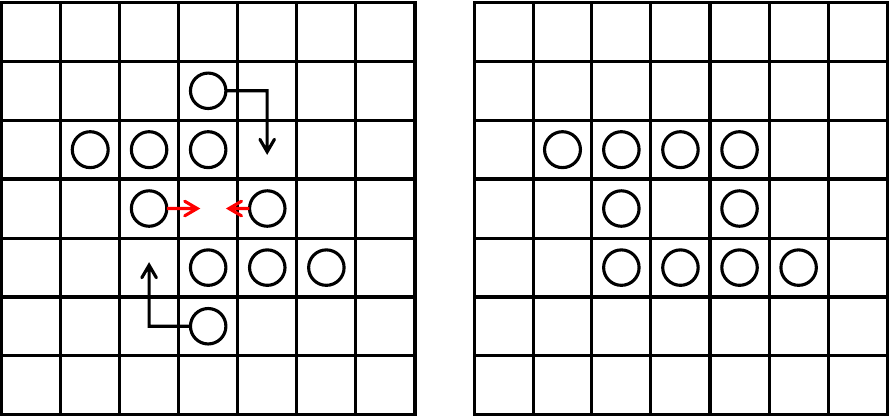} \\
    $(c)$ 
   \end{minipage} &
   \begin{minipage}[t]{0.45\hsize}
    \centering
    \includegraphics[height=1.5cm]{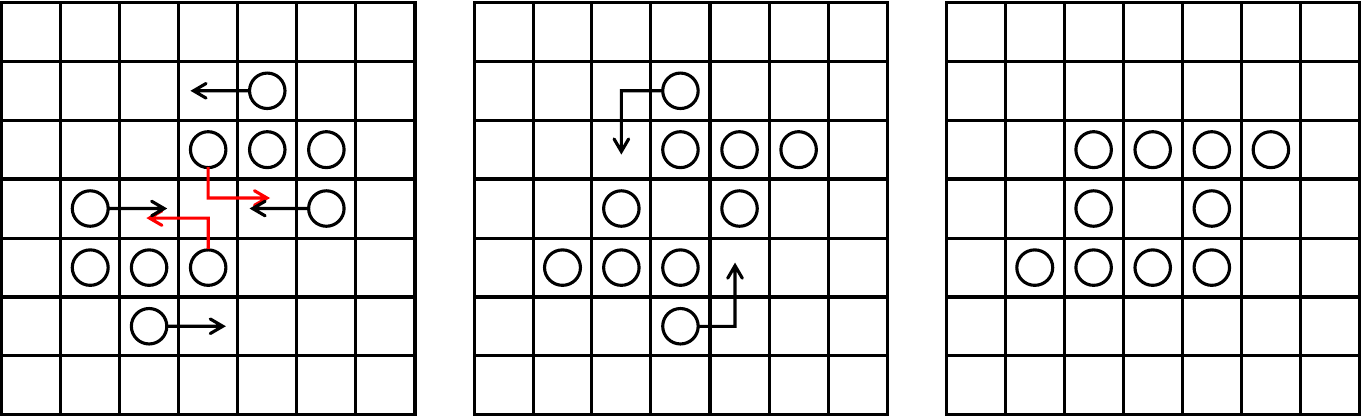} \\
    $(d)$ 
   \end{minipage}
   \vspace{3mm} \\ 
      \begin{minipage}[t]{0.45\hsize}
    \centering
    \includegraphics[height=1.5cm]{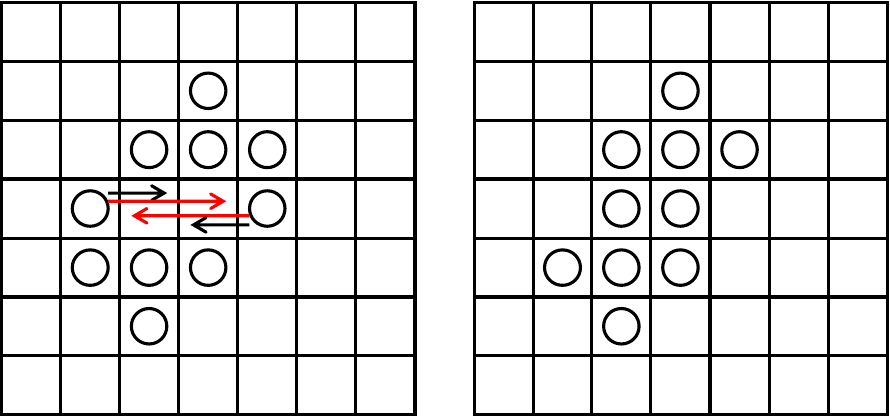} \\ 
    $(e)$ 
   \end{minipage} &
   \begin{minipage}[t]{0.45\hsize}
    \centering
    \includegraphics[height=1.5cm]{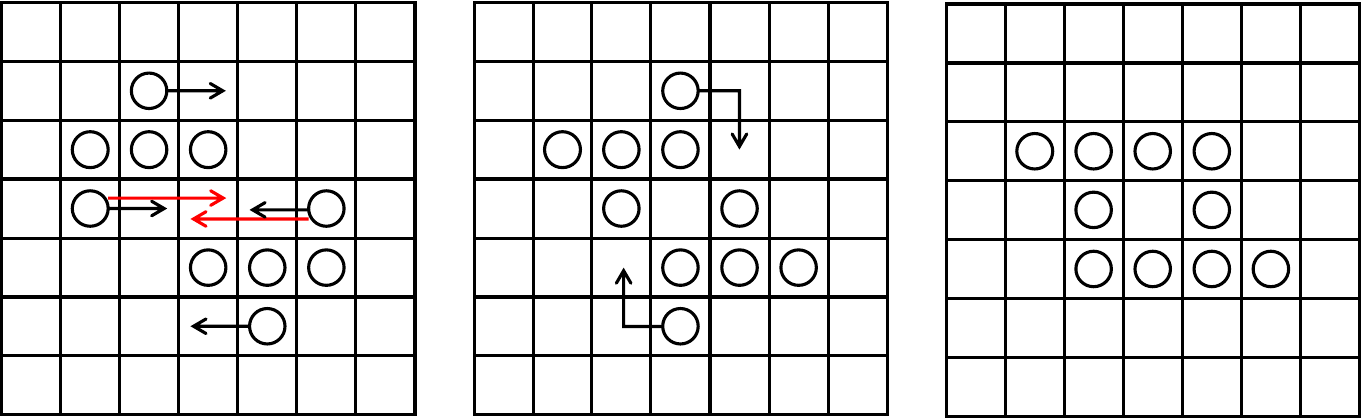} \\ 
    $(f)$ 
   \end{minipage}
   \vspace{3mm} \\ 
   \begin{minipage}[t]{0.45\hsize} 
   \centering
    \includegraphics[height=1.5cm]{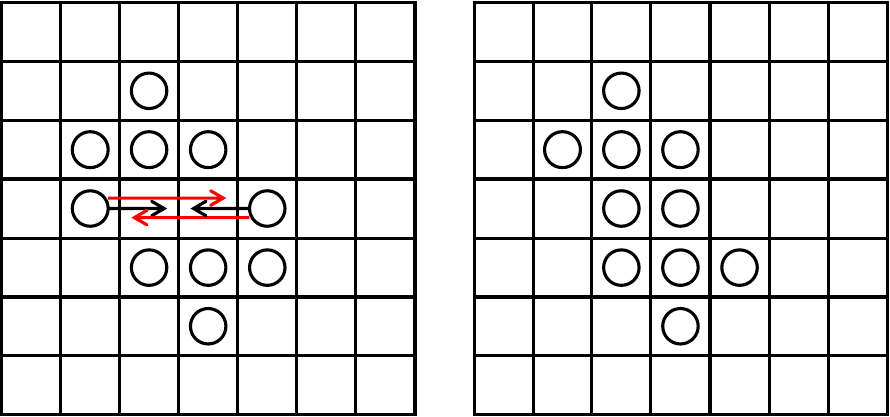} \\ 
    $(g)$ 
   \end{minipage} &
   \begin{minipage}[t]{0.45\hsize}
    \centering
    \includegraphics[height=1.5cm]{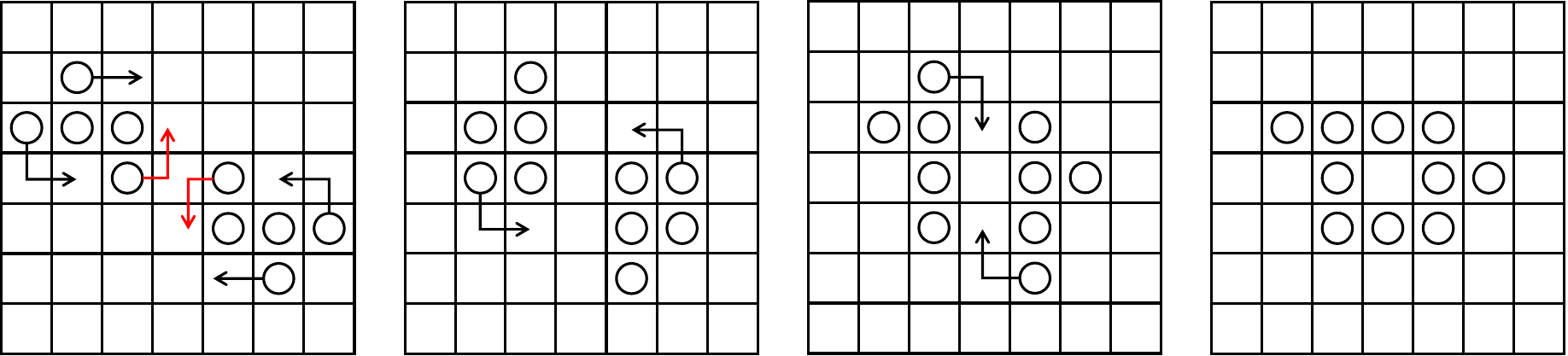} \\ 
    $(h)$ 
   \end{minipage}
   \vspace{3mm} \\ 
   \begin{minipage}[t]{0.45\hsize}
    \centering
    \includegraphics[height=1.5cm]{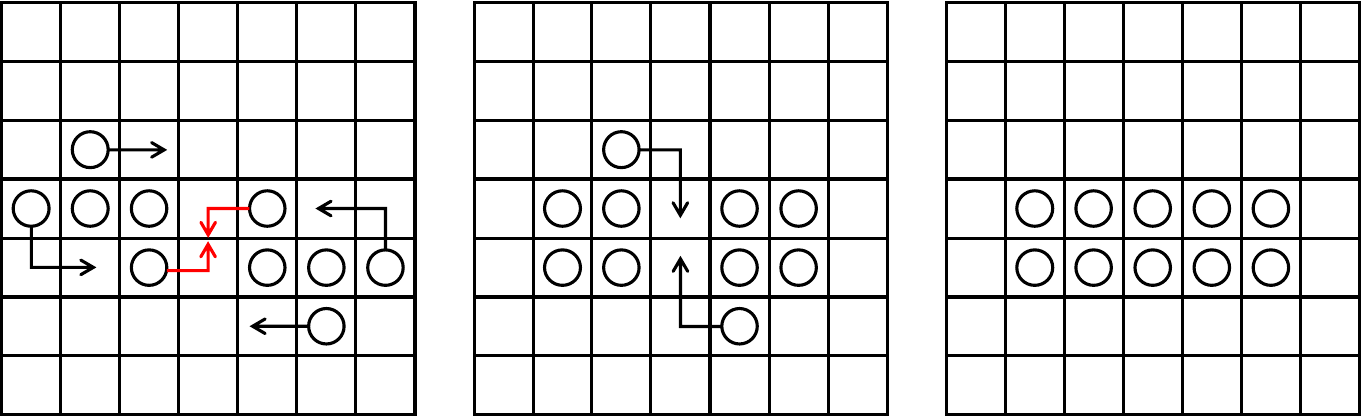} \\ 
    $(i)$
   \end{minipage} &
   \begin{minipage}[t]{0.45\hsize}
    \centering
    \includegraphics[height=1.5cm]{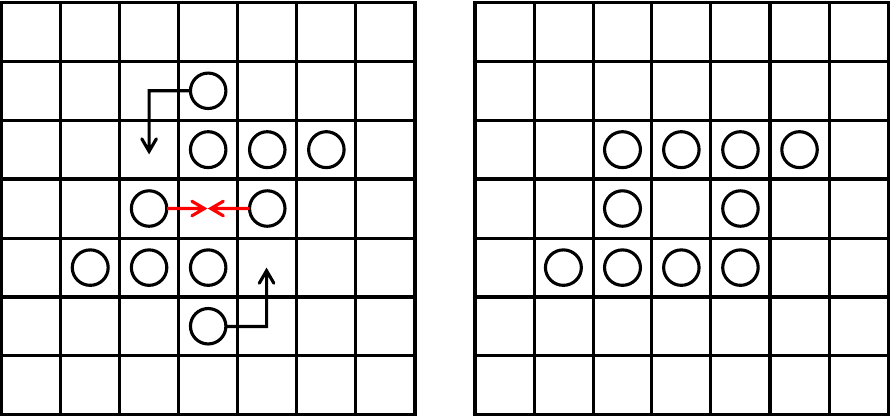} \\
    $(j)$ 
   \end{minipage}
   \vspace{3mm} \\ 
   \multicolumn{2}{c}{
    \begin{minipage}[t]{0.9\hsize}
     \centering
     \includegraphics[height=1.5cm]{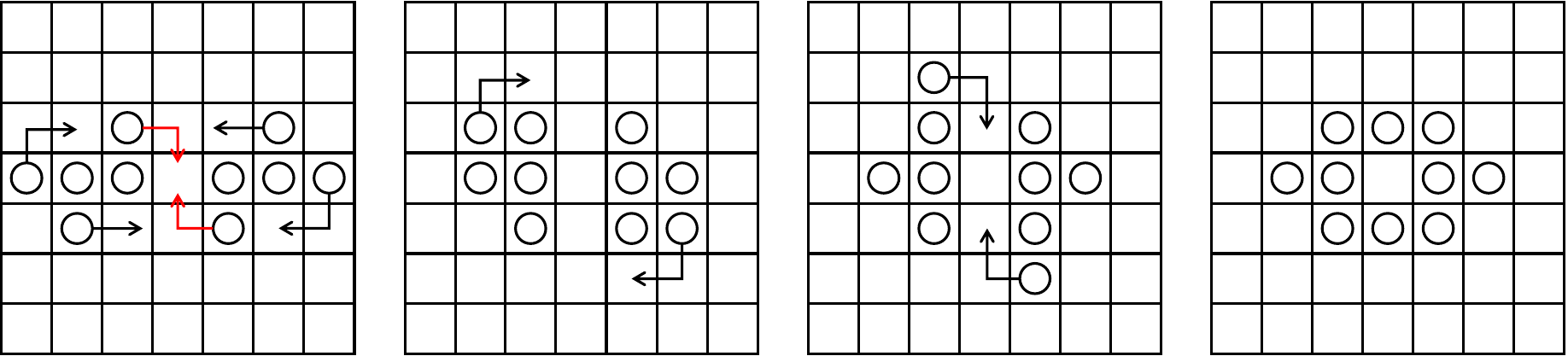} \\ 
     $(g)$ 
     \end{minipage}
   }
   \vspace{3mm} \\ 
   \multicolumn{2}{c}{
    \begin{minipage}[t]{0.9\hsize}
    \centering
    \includegraphics[height=1.5cm]{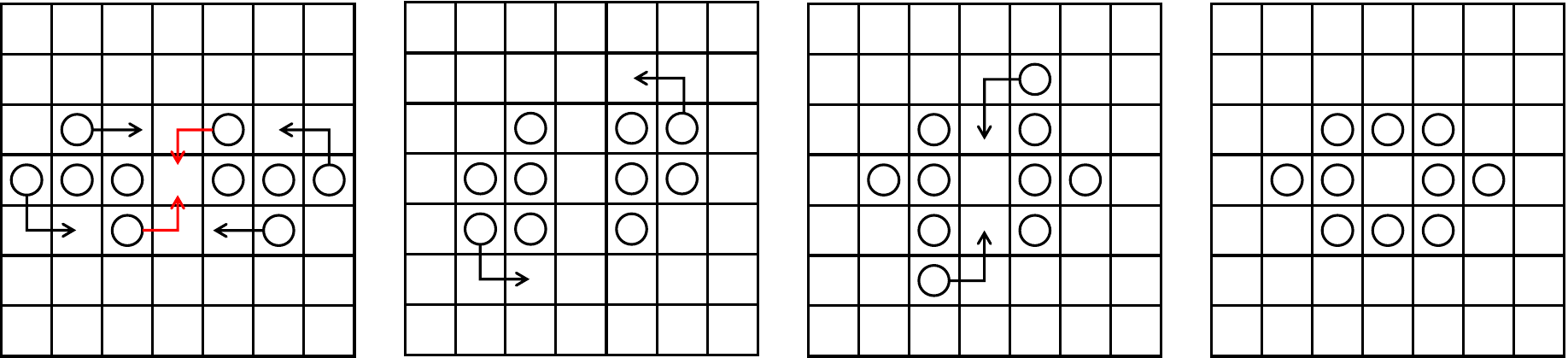} \\
    $(h)$ 
    \end{minipage}
   }
  \end{tabular}
  \caption{Exceptional movements for symmetric configurations. The red arrows show the movements that cannot be performed due to collisions. The black arrows show the exceptional movements.} 
  \label{fig:merge-ex} 
 \end{figure}

We then consider symmetric initial configurations, where the two MRSs have the same view. 
In this case, the proposed algorithm makes the two MRSs to perform 
the transformations shown in Figure~\ref{fig:Dec_TD} until their states become 
$S^5_1$ or $S^5_2$. 
The resulting configuration is also symmetric and the two MRSs are in the same state. 
Then, the proposed algorithm makes the two MRSs to follow the same procedure 
as $R_1$ in the asymmetric initial configurations. 
They check the overlap of their smallest enclosing rectangles and 
perform $M_{x+}$ or $M_{x-}$ until they become $x$-overlapping. 
Then, they turn by $T_{S^5_1}$ or $T_{S^5_2}$ and 
perform $M_{x+}$ or $M_{x-}$ until they merge. 
All these moves are symmetric with respect to the center of symmetry 
in the initial configuration. 
However, in the final step, the two MRS may caught in a deadlock 
due to collisions of modules. 
We incorporate exceptional moves shown in Figure~\ref{fig:merge-ex} 
for these cases to the proposed algorithm. 
Note that in an execution of the proposed merge algorithm, 
if the two MRS performs a turn ($T_{S^5_1}$ or $T_{S^5_2}$), 
their smallest enclosing rectangles overlap by one row or by two rows after the turn 
due to symmetric movement of the two MRSs. 
In Figure~\ref{fig:merge-ex}, the first three cases $(a), (b)$ and $(c)$ show 
exceptional moves for collisions in $M_{x+}$ with $S^5_1$. 
The red arrows show the movements of $M_{x+}$ that cannot be taken due to collisions. 
The proposed algorithm makes the modules to perform movements shown with 
black arrows to avoid the movements shown in red arrows. 
Other movements of $M_{x+}$ with $S^5_1$ does not cause collisions. 
In Figure~\ref{fig:merge-ex}, the next two cases $(d)$ and $(e)$ show 
exceptional movements for collisions in $M_{x+}$ with $S^5_2$. 
These two cases show all possible collisions of $M_{x+}$ with $S^5_2$. 
In Figure~\ref{fig:merge-ex}, the next two cases $(f)$ and $(g)$ show 
exceptional movements for collisions in $M_{x-}$ with $S^5_1$. 
These two cases show all possible collisions of $M_{x-}$ with $S^5_1$. 
In Figure~\ref{fig:merge-ex}, the next three cases $(h), (i)$ and $(j)$ show 
exceptional movements for collisions in $M_{x-}$ with $S^5_2$. 
In Figure~\ref{fig:merge-ex}, the next two cases $(d)$ and $(e)$ show 
In Figure~\ref{fig:merge-ex}, the last two cases $(k)$ and $(l)$ show 
exceptional movements when the two MRS are in the same three rows, 
which may occur in an initial configuration. 
These two cases show all possible collisions for this case. 

\clearpage

\begin{figure}[t]
 \centering
 \includegraphics[width=2.5cm]{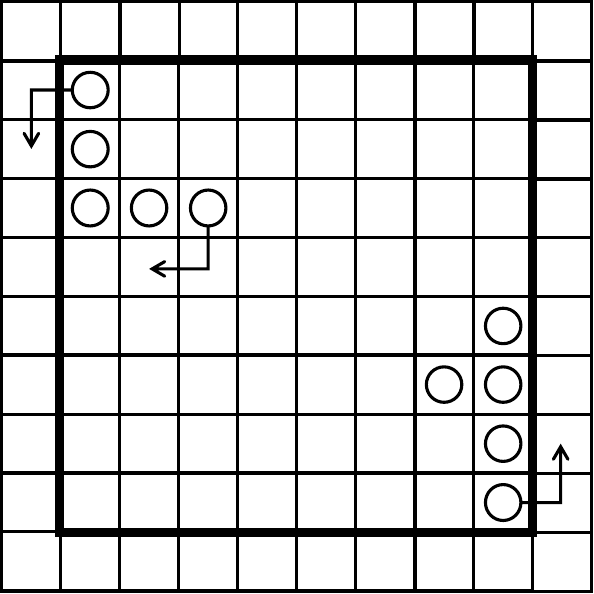} 
 \caption{MRSs go outside of the visibility range of some modules during the execution of the merge algorithm when visibility size is $7$.} 
 \label{fig:merge_k7} 
\end{figure}

In the proposed merge algorithm, we assume visibility range of $9$. 
Consider a configuration shown in Figure~\ref{fig:merge_k7}, 
where the smallest enclosing rectangle of the two MRSs is a 
regular square of size $8$. 
Hence, each module can observe all the ten modules even when the visibility range is $7$. 
Let $R_1$ and $R_2$ be the two MRSs. 
When $R_1$ is in state $S^5_{11}$ and $R_2$ is in state $S^5_{8}$, 
$R_1$ performs the transformation to $S^5_{18}$ shown in Figure~\ref{fig:Inc_TD} 
and $R_2$ performs the transformation to $S^5_1$ shown in Figure~\ref{fig:Dec_TD}. 
However, during the translations, some modules of $R_1$ and $R_2$ 
go outside of the regular square of size $8$.
Thus, if the visibility range is $7$, 
there exists some module that cannot observe some of the ten modules. 
These modules cannot continue the merge algorithm. 
In the proposed algorithm, 
when an MRS executes transformations to $S^5_1, S^5_2$ and $S^5_{18}$, 
it can go outside of its initial smallest enclosing rectangle at most 
once and at most by one. 
In the transformations to $S^5_{18}$ shown in Figure~\ref{fig:Inc_TD}, 
these cases are the moves starting from 
$S^5_{3}, S^5_{4}, S^5_{5}, S^5_{6}, S^5_{7}, S^5_{8}, S^5_{9}, S^5_{11}, S^5_{13}, S^5_{14}, S^5_{15}$, 
and in the transformations to $S^5_1$ or $S^5_2$ shown in Figure~\ref{fig:Dec_TD}, 
these cases are the moves starting from 
$S^5_{3}, S^5_{4}, S^5_{7}, S^5_{8}, S^5_{15}$. 
Then, in the worst case, visibility size of $9$ is sufficient to keep the 
modules visible to each other and continue the proposed merge algorithm. 

We show the pseudocode of the proposed algorithm in 
Algorithm~\ref{alg:procedure} and Algorithm~\ref{alg:merge}. 

\begin{algorithm}[t]
\caption{$\textsc{Horizontal\_movement}(m)$ for module $m$} 
\label{alg:procedure} 
\begin{tabbing}
xxx \= xxx \= xxx \= xxx \= xxx \= xxx \= xxx \= xxx \= xxx \= xxx
\kill 
{\bf Notation} \\ 
\> $R$: MRS that module $m$ belongs to \\
\> $R'$: the other MRS \\ 
\\ 
{\bf Precondition} \\ 
\> $R$ is in state $S^5_1$ or $S^5_2$ \\ 
{\bf Algorithm} \\ 
{\bf If} $R'$ is not $x$-overlapping with $R$ {\bf then} \\ 
\> {\bf If} the position of the two MRSs is one of those shown in Figure~\ref{fig:merge-ex} {\bf then} \\  
\> \> // Two MRSs are symmetric \\ 
\> \> Perform the movements shown in Figure~\ref{fig:merge-ex} \\  
\> {\bf Else} \\
\> \> {\bf If} $R'$ is in the positive $x$ direction of $R$'s local coordinate system {\bf then} \\  
\> \> \> $M_{x+}$ \\ 
\> \> {\bf Else} \\ 
\> \> \> $M_{x-}$ \\
\> \> {\bf Endif} \\ 
\> {\bf Endif} \\ 
{\bf Else} // $R$ and $R'$ are $x$-overlapping \\ 
\> {\bf If} $R$ is in state $S^5_1$ {\bf then} $T_{S^5_1}$ \\ 
\> {\bf If} $R$ is in state $S^5_2$ {\bf then} $T_{S^5_2}$ \\ 
{\bf Endif} \\
\end{tabbing}
\end{algorithm}

\begin{algorithm}[ht]
\caption{Merge algorithm for module $m$} 
\label{alg:merge} 
\begin{tabbing}
xxx \= xxx \= xxx \= xxx \= xxx \= xxx \= xxx \= xxx \= xxx \= xxx
\kill 
{\bf Notation} \\ 
\> $R$: MRS that module $m$ belongs to \\ 
\> $R'$: the other MRS \\ 
\\ 
{\bf Algorithm} \\ 
\> {\bf If} $R$ and $R'$ have different labels {\bf then} \\ 
\> \> {\bf If} $R$ has the larger label and $R$ is not in $S_{18}^5$ {\bf then} \\ 
\> \> \> Execute the corresponding movement in Figure~\ref{fig:Inc_TD}\\ 
\> \> {\bf Else} // $R$ has the smaller label \\ 
\> \> \> {\bf If} $R'$ is in $S^5_{18}$ {\bf then} \\ 
\> \> \> \> {\bf If} $R$ is not in $S^5_1$ nor $S^5_2$ {\bf then} \\ 
\> \> \> \> \> Execute the corresponding movement in Figure~\ref{fig:Dec_TD}\\ 
\> \> \> \> {\bf Else} \\
\> \> \> \> \> $\textsc{Horizontal\_movement}(m)$ \\ 
\> \> \> \> {\bf Endif} \\ 
\> \> \> {\bf Endif} \\ 
\> \> {\bf Endif} \\ 
\> {\bf Else} // $R$ and $R'$ have the same label \\ 
\> \> {\bf If} the views of $R$ and $R'$ are different {\bf then} \\ 
\> \> \> {\bf If} $R$ has the larger view and $R$ is not in $S_{18}^5$ {\bf then} \\ 
\> \> \> \> Execute the corresponding movement in Figure~\ref{fig:Inc_TD}\\ 
\> \> \> {\bf Else} // $R$ has the smaller view \\ 
\> \> \> \> {\bf If} $R'$ is in $S^5_{18}$ {\bf then} \\ 
\> \> \> \> \> {\bf If} $R$ is not in $S_{1}^5$ nor $S^5_2$ {\bf then} \\ 
\> \> \> \> \> \> Execute the corresponding movement in Figure~\ref{fig:Dec_TD}\\ 
\> \> \> \> \> {\bf Endif} \\ 
\> \> \> \> {\bf Endif} \\ 
\> \> \> {\bf Else} // $R$ and $R'$ are symmetric \\ 
\> \> \> \> {\bf If} $R$'s is not in $S^5_{1}$ nor $S^5_2$ {\bf then} \\ 
\> \> \> \> \> Execute the corresponding movement in Fig~.\ref{fig:Dec_TD}\\ 
\> \> \> \> {\bf Else} \\ 
\> \> \> \> $\textsc{Horizontal\_movement}(m)$ \\ 
\> \> \> {\bf Endif} \\
\> \> {\bf Endif} \\
\> {\bf Endif} \\
\end{tabbing}
\end{algorithm}

\begin{theorem} 
Two MRSs each of which consists of five modules not equipped with the global compass 
can solve the merge problem 
from an arbitrary initial configuration, where each module of the two MRSs 
can observe all the modules in its visibility. 
\end{theorem}

We finally show how to synthesize the rendezvous algorithm and the merge algorithm. 
Each module starts the rendezvous algorithm and switches to the merge algorithm 
as soon as the smallest enclosing rectangle of the two MRSs becomes 
a regular square of size $8 \times 8$ or a smaller square. 
The required visibility range is $9$.

\section{Impossibility} 
\label{sec:impossibility}

We finally show necessity of the five modules to solve the rendezvous problem. 
In~\cite{DYKY22}, Doi et al. showed that an MRS that consists of four modules 
not equipped with the global compass can move horizontally or vertically. 
By considering movements on walls, an MRS can move on a cycle formed by 
two horizontal (or vertical) lines connected by two constant-length 
lines on the walls. 
When these cycles for the two MRSs do not overlap and the modules of one MRS 
cannot observe the other MRS, the two MRSs never meet. 
The authors also showed that when each MRS consists of less than four modules, 
the MRS cannot move. 
Thus, we have the following theorem. 

\begin{theorem}
Two MRSs each of which consists of less than five modules not equipped with the global compass 
cannot solve the rendezvous problem. 
\end{theorem}

\section{Conclusion} 
\label{sec:concl}

In this paper we presented a rendezvous algorithm and a merge algorithm 
for two MRSs, each of which consists of five modules 
not equipped with the global compass. 
We then showed that five modules for each MRS are necessary to solve the rendezvous problem. 
Due to the page limitation, we omit the fact that the merge problem can be solved 
by two MRSs of size four. 

While existing results consider distributed coordination among the modules 
so that they collectively form a single MRS, 
we considered distributed coordination among multiple MRSs. 
We believe the results open up new vistas for distributed coordination theory 
for MRSs. 
There are many future directions including 
conventional distributed coordination problems such as leader election, 
synchronization, and fault tolerance, 
and mobile computing problems such as gathering, separation, and patrolling. 
It is also important to develop basic buildingblocks for distributed algorithms for MRSs instead of tailor-made algorithms.

\clearpage 

\bibliographystyle{splncs04}
\bibliography{papers} 

\end{document}